\documentclass[dvipdfmx,10pt]{article}

\usepackage{amsfonts}
\usepackage{amsmath}
\usepackage{bm}
\usepackage{myDefinition2}
\usepackage{graphicx}
\usepackage{subfigure} 
\usepackage{algorithm}
\usepackage{algorithmic}

\def\Vec#1{\mbox{\boldmath $#1$}}

\newcommand{\argmin}{\mathop{\rm argmin}\limits}

\usepackage[margin=1in]{geometry}

\title{Homotopy Continuation Approaches\\for Robust SV Classification and Regression}

\date{September 12, 2015}

\author{
S.~Suzumura\\
Nagoya Institute of Technology\\
\texttt{suzumura.mllab.nit@gmail.com}
\\
\and
K.~Ogawa\\
Nagoya Institute of Technology\\
\texttt{ogawa.mllab.nit@gmail.com}
\\
\and
M.~Sugiyama\\
The University of Tokyo\\
\texttt{sugi@k.u-tokyo.ac.jp}
\\
\and
M.~Karasuyama\\
Nagoya Institute of Technology\\
\texttt{karasuyama@nitech.ac.jp}
\\
\and
I.~Takeuchi\thanks{Corresponding Author}\\
Nagoya Institute of Technology\\
\texttt{takeuchi.ichiro@nitech.ac.jp}
}

\begin{document}

\maketitle

\begin{abstract}
In support vector machine (SVM) applications 
with unreliable data
that contains a portion of outliers, 
non-robustness of SVMs often causes considerable performance deterioration.
Although
many approaches for improving the robustness of SVMs have been studied, 
two major challenges remain in robust SVM learning. 
First,
robust learning algorithms
are essentially formulated
as non-convex optimization problems
because the loss function must be designed
to alleviate the influence of outliers.
It is thus important to develop a non-convex optimization method
for robust SVM 
that can find a good local optimal solution. 
The second practical issue is how
one can tune the hyperparameter that 
controls the balance between robustness and efficiency. 
Unfortunately, 
due to the non-convexity, 
robust SVM solutions with slightly different hyper-parameter values can be significantly different,
which makes model selection highly unstable. 
In this paper,
we address these two issues simultaneously 
by introducing a novel \emph{homotopy} approach to non-convex robust SVM learning. 
Our basic idea is to introduce parametrized formulations of robust SVM 
which 
bridge 
the standard SVM
and
fully robust SVM
via the parameter that represents the influence of outliers.
We characterize the necessary and sufficient conditions of the local optimal solutions of robust SVM,
and
develop an algorithm 
that can trace a path of local optimal solutions when the influence of outliers is gradually decreased.
An advantage of our homotopy approach is that
it can be interpreted as simulated annealing,
a common approach for finding a good local optimal solution in non-convex optimization problems. 
In addition,
our homotopy method allows stable and efficient model selection based on the path of local optimal solutions.
Empirical performances of the proposed approach are demonstrated
through intensive numerical experiments
both on robust classification and regression problems. 

\end{abstract}

\section{Introduction}
\label{sec:introduction}
\begin{figure}[!h]
  \begin{tabular}{cc}
  \begin{minipage}{0.5\hsize}
    \subfigure[Standard SVC] {
      \includegraphics[width=0.7\linewidth]{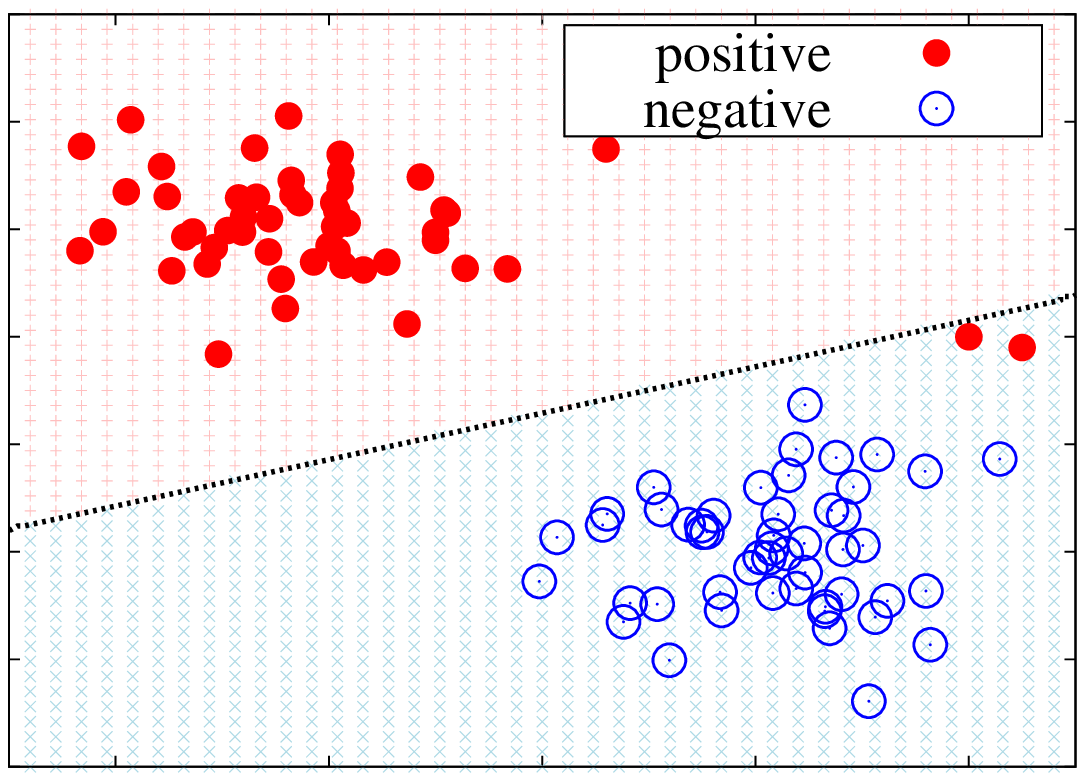}
      \label{fig:cmp_SVM_with_RSVM_a}
    }
  \end{minipage}
  \begin{minipage}{0.5\hsize}
    \subfigure[Robust SVC] {
      \includegraphics[width=0.7\linewidth]{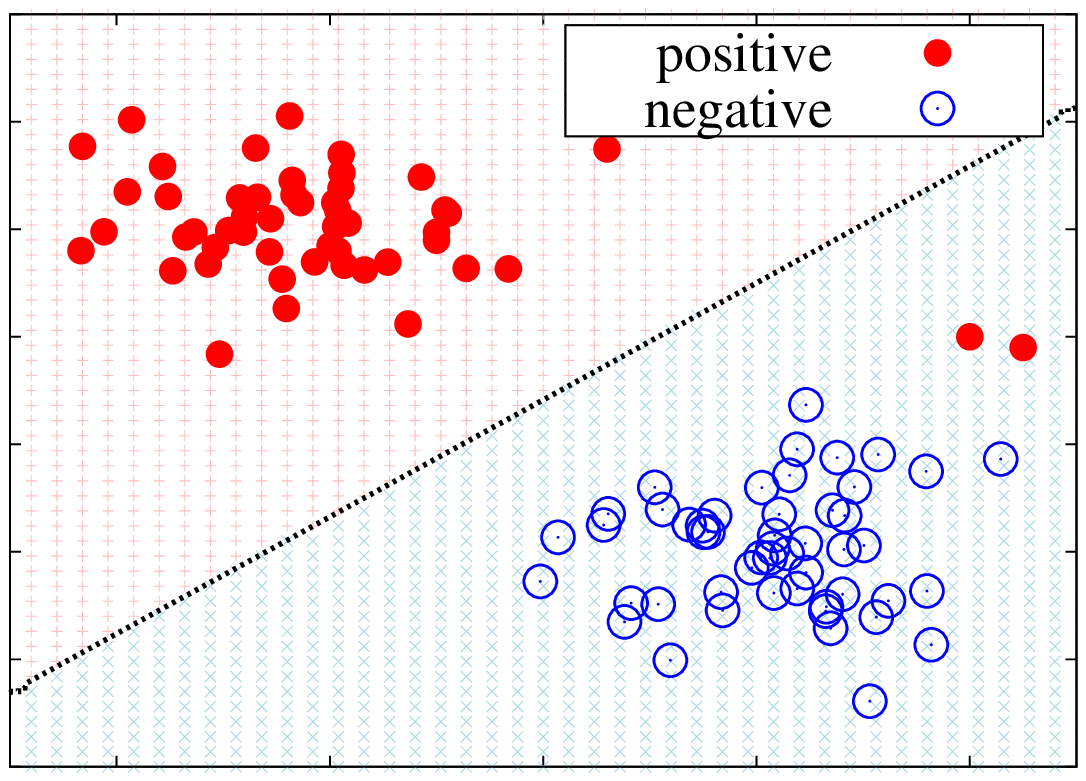}
      \label{fig:cmp_SVM_with_RSVM_b}
    }
  \end{minipage}
  \end{tabular}
  \begin{tabular}{cc}
  \begin{minipage}{0.5\hsize}
    \subfigure[Standard SVR] {
      \includegraphics[width=0.7\linewidth]{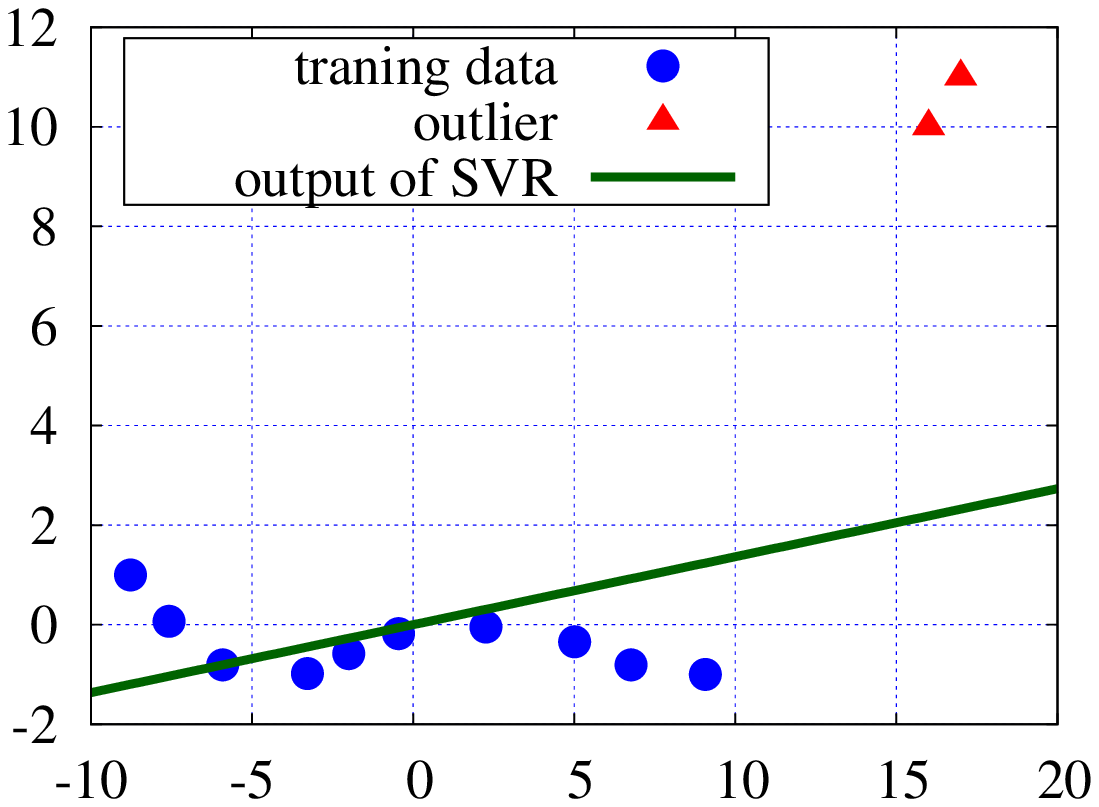}
      \label{fig:example_csvr}
    }
  \end{minipage}
  \begin{minipage}{0.5\hsize}
    \subfigure[Robust SVR] {
      \includegraphics[width=0.7\linewidth]{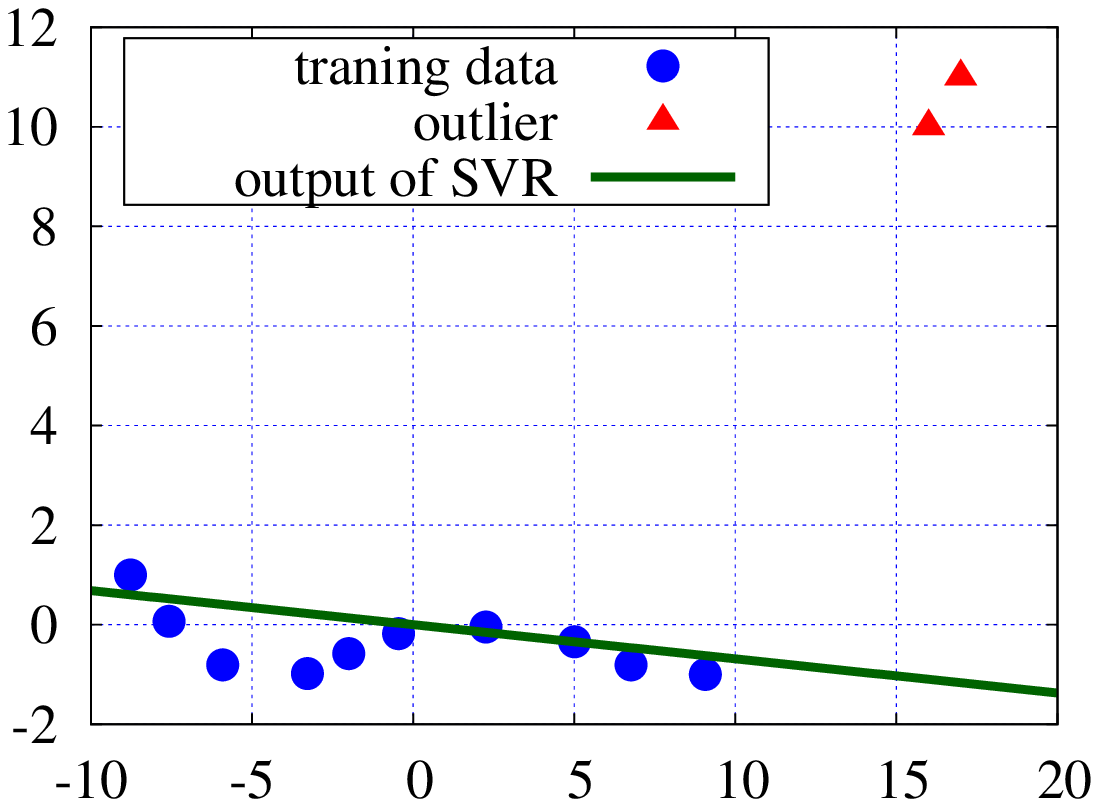}
      \label{fig:example_rsvr}
    }
  \end{minipage}
  \end{tabular}
  \caption{Illustrative examples of (a) standard SVC (classification) and (b) robust SVC, (c) standard SVR (regression) and (d) robust SVR on toy dataset.
 In robust SVM, the classification and regression results are not sensitive to the two red outliers in the right-hand side of the plots.}
  \label{fig:cmp_SVM_with_RSVM}
\end{figure}

The
\emph{support vector machine} (SVM) 
has been one of the most successful machine learning algorithms
\cite{Boser92a,Cortes95a,Vapnik98a}.
However, 
in recent practical machine learning applications with less reliable data
that contains a portion of outliers 
(e.g.,
consider situations where
the labels are automatically obtained
by semi-supervised learning \cite{zhu2003semi,zhu2005semi}
or
manually annotated in crowdsourcing framework \cite{yuen2011survey,mao2015survey}),
non-robustness of the SVM 
often causes considerable performance deterioration.
See
\figurename \ref{fig:cmp_SVM_with_RSVM} 
for examples of robust classification and regression.
\subsection{Limitations of Existing Robust Classification and Regression}
Although a great deal of efforts have been devoted
to improving the robustness of SVM and other similar learning algorithms
\cite{mason00a,Shen03,krause04a,Liu05a,Liu06a,Xu06,Collobert06a,wu07a,masnadi-shirazi08a,Freund09,Yu10},
the problem of learning SVM under massive noise still poses two major challenges.

\begin{figure*}[!h]
 \begin{center}
  \begin{tabular}{ccccc}
   \includegraphics[width=0.17\textwidth]{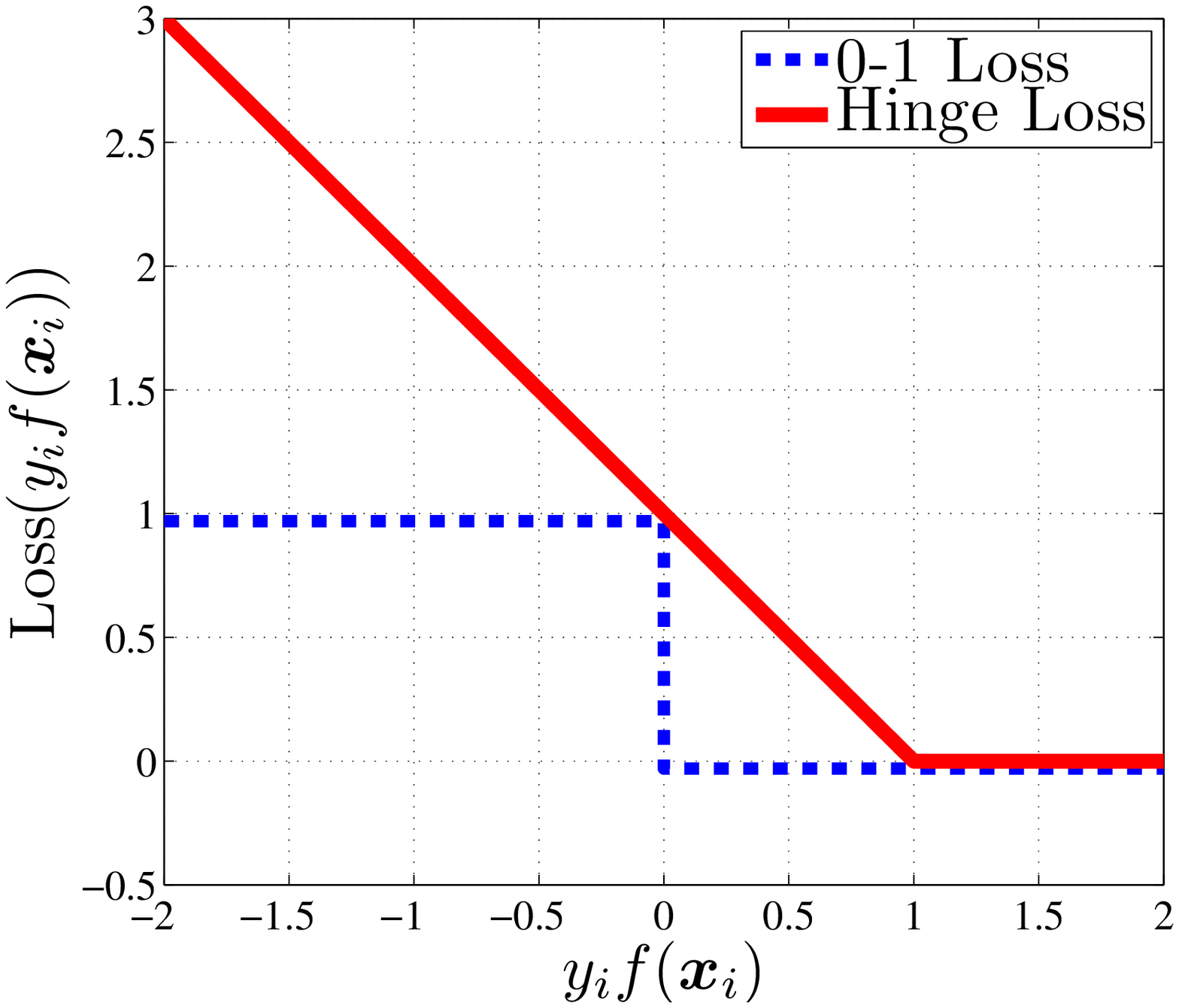} &
   \includegraphics[width=0.17\textwidth]{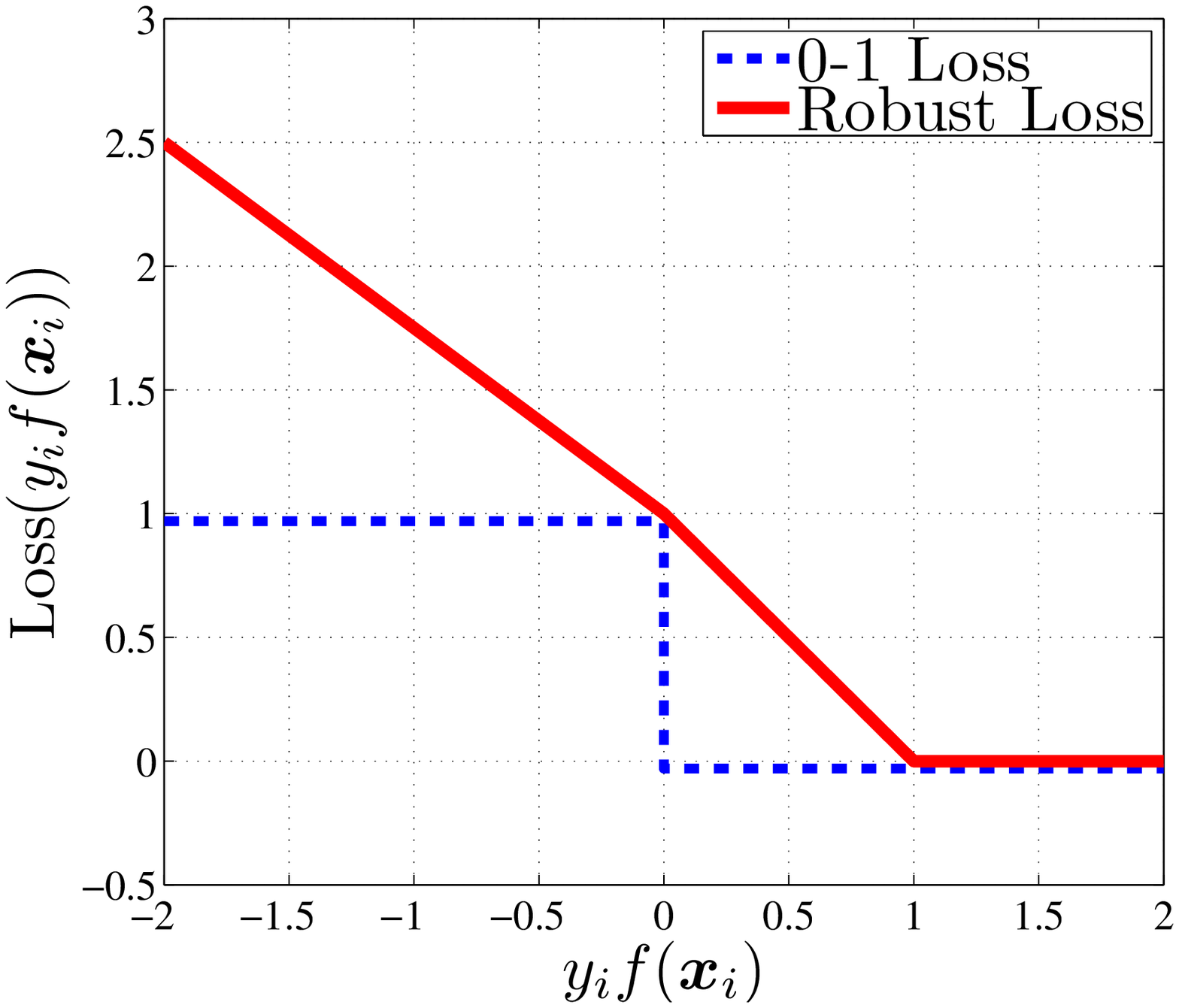} &
   \includegraphics[width=0.17\textwidth]{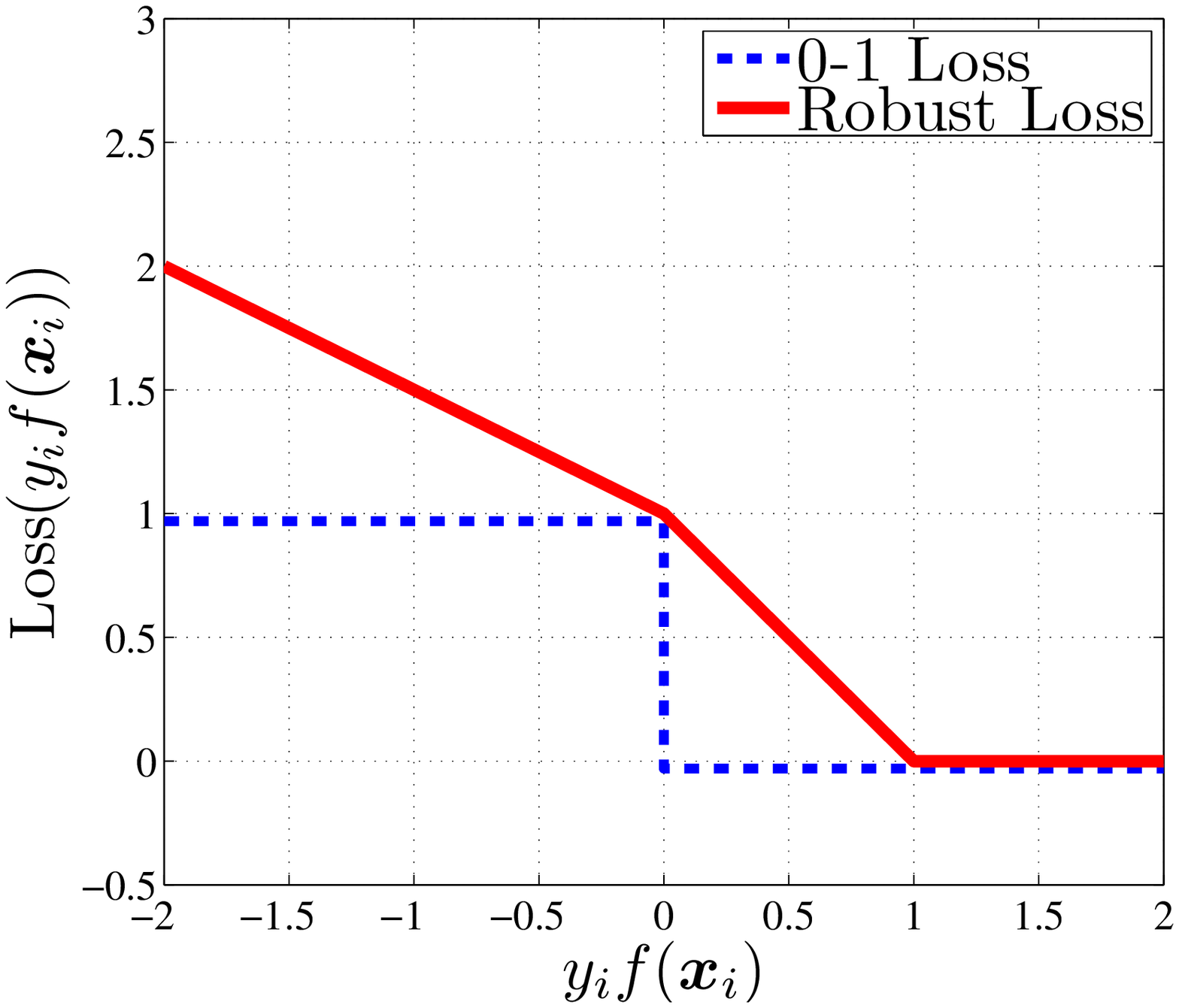} &
   \includegraphics[width=0.17\textwidth]{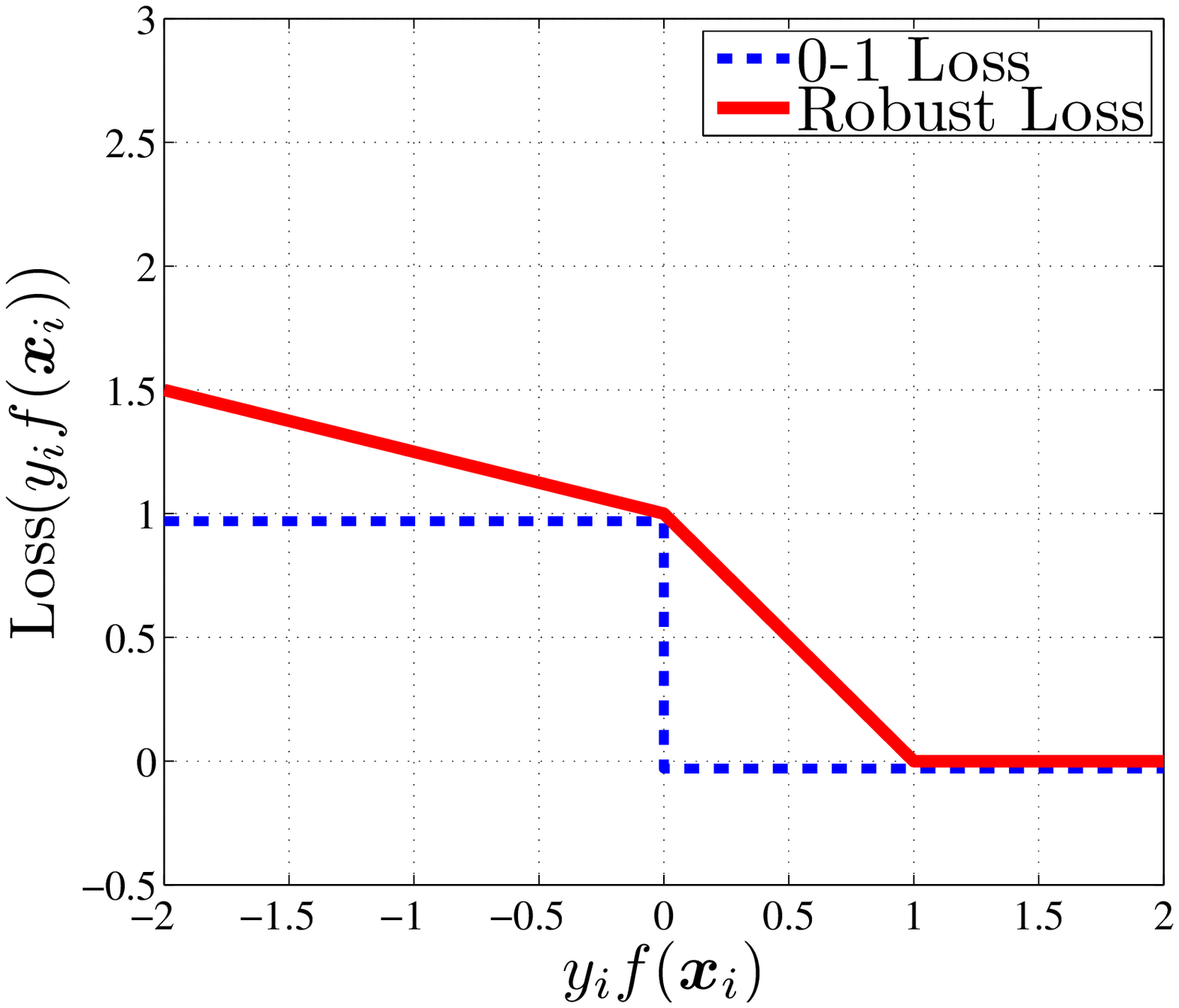} &
   \includegraphics[width=0.17\textwidth]{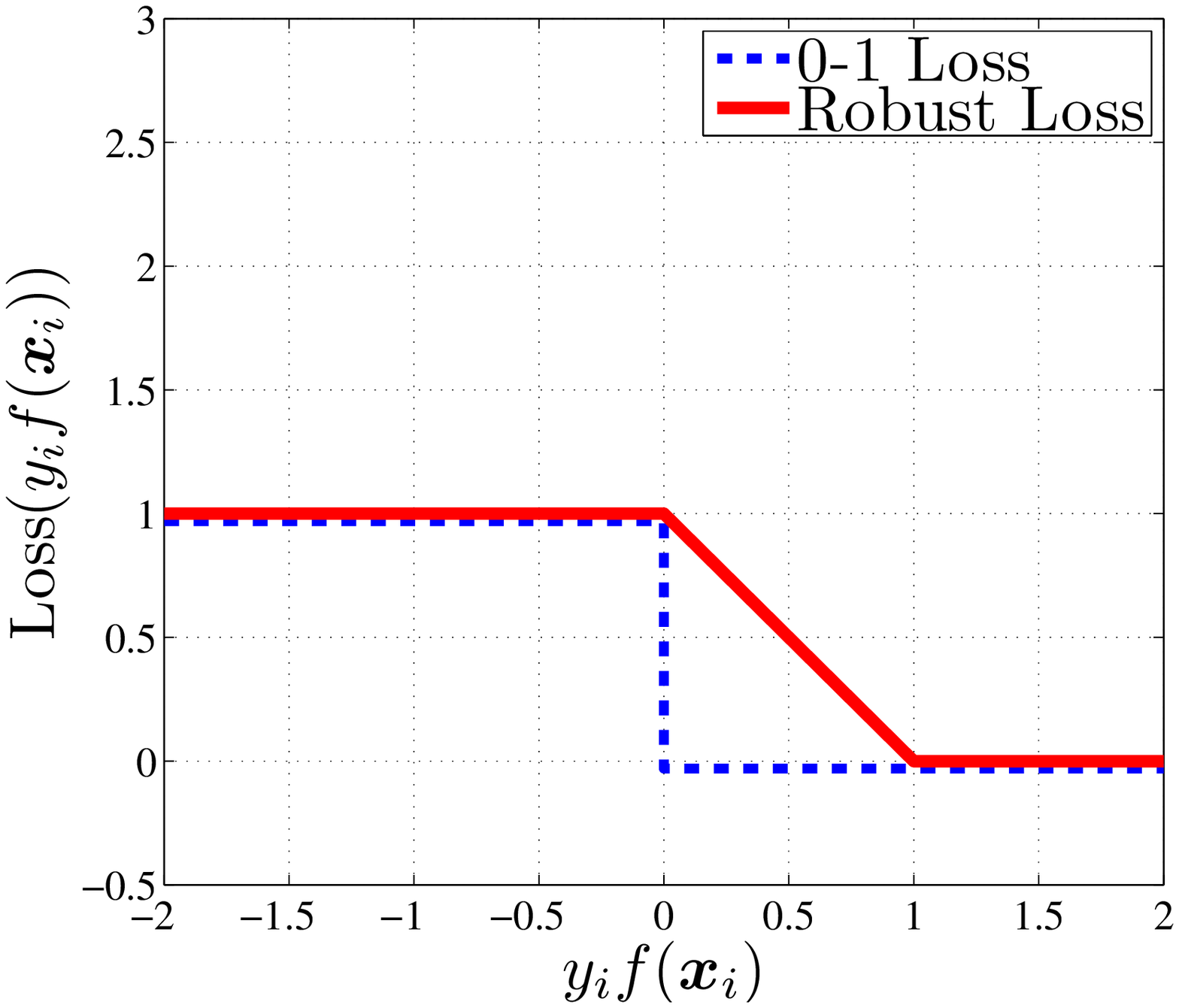} \\
   \footnotesize{$\theta = 1$, $s = 0$} &
   \footnotesize{$\theta = 0.75$, $s = 0$} &
   \footnotesize{$\theta = 0.5$, $s = 0$} &
   \footnotesize{$\theta = 0.25$, $s = 0$} &
   \footnotesize{$\theta = 0$, $s = 0$} \\		   
   \multicolumn{5}{c}{\small (a) Homotopy computation with decreasing $\theta$ from 1 to 0.} \\
   \includegraphics[width=0.17\textwidth]{loss_hinge} &
   \includegraphics[width=0.17\textwidth]{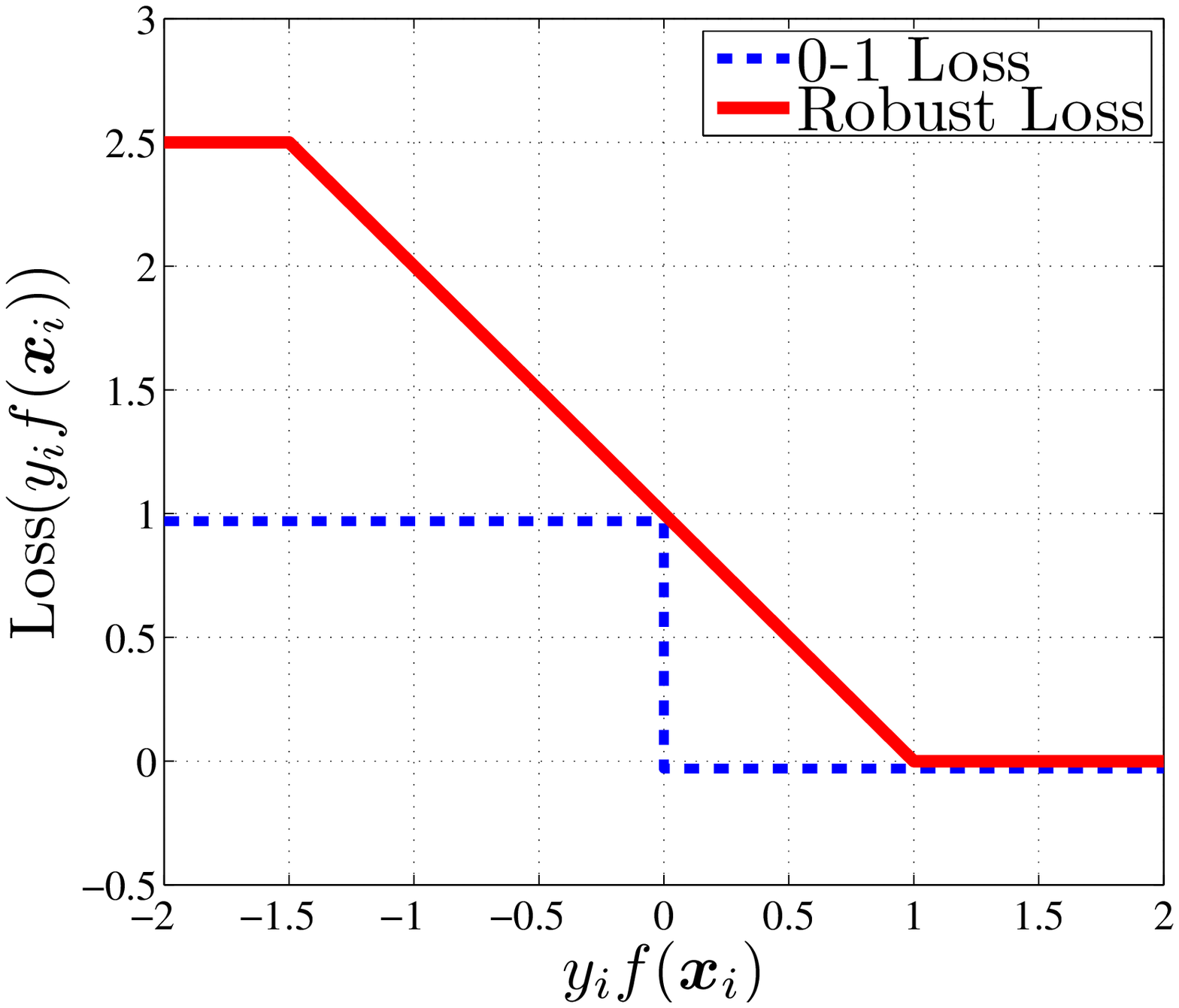} &
   \includegraphics[width=0.17\textwidth]{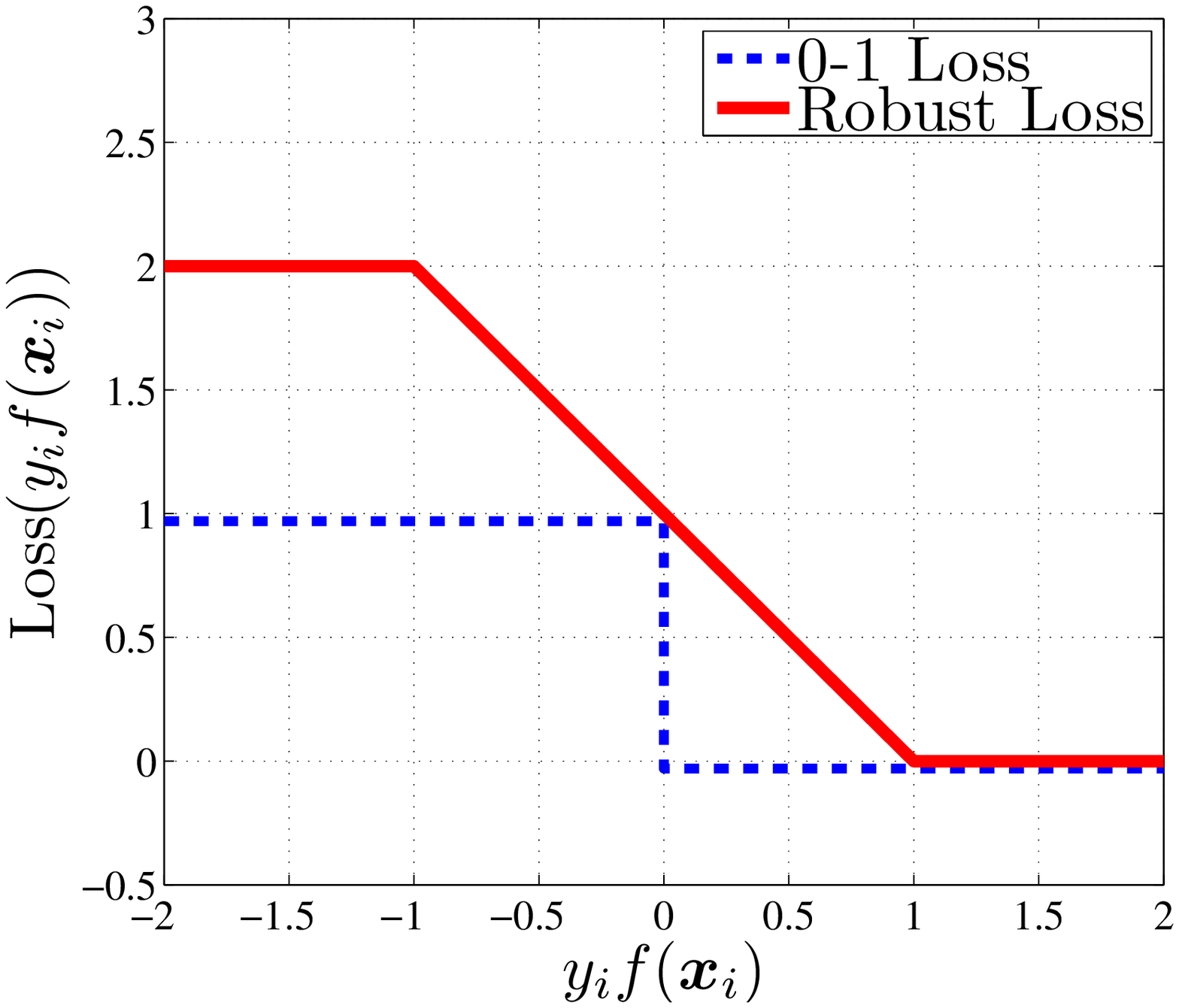} &
   \includegraphics[width=0.17\textwidth]{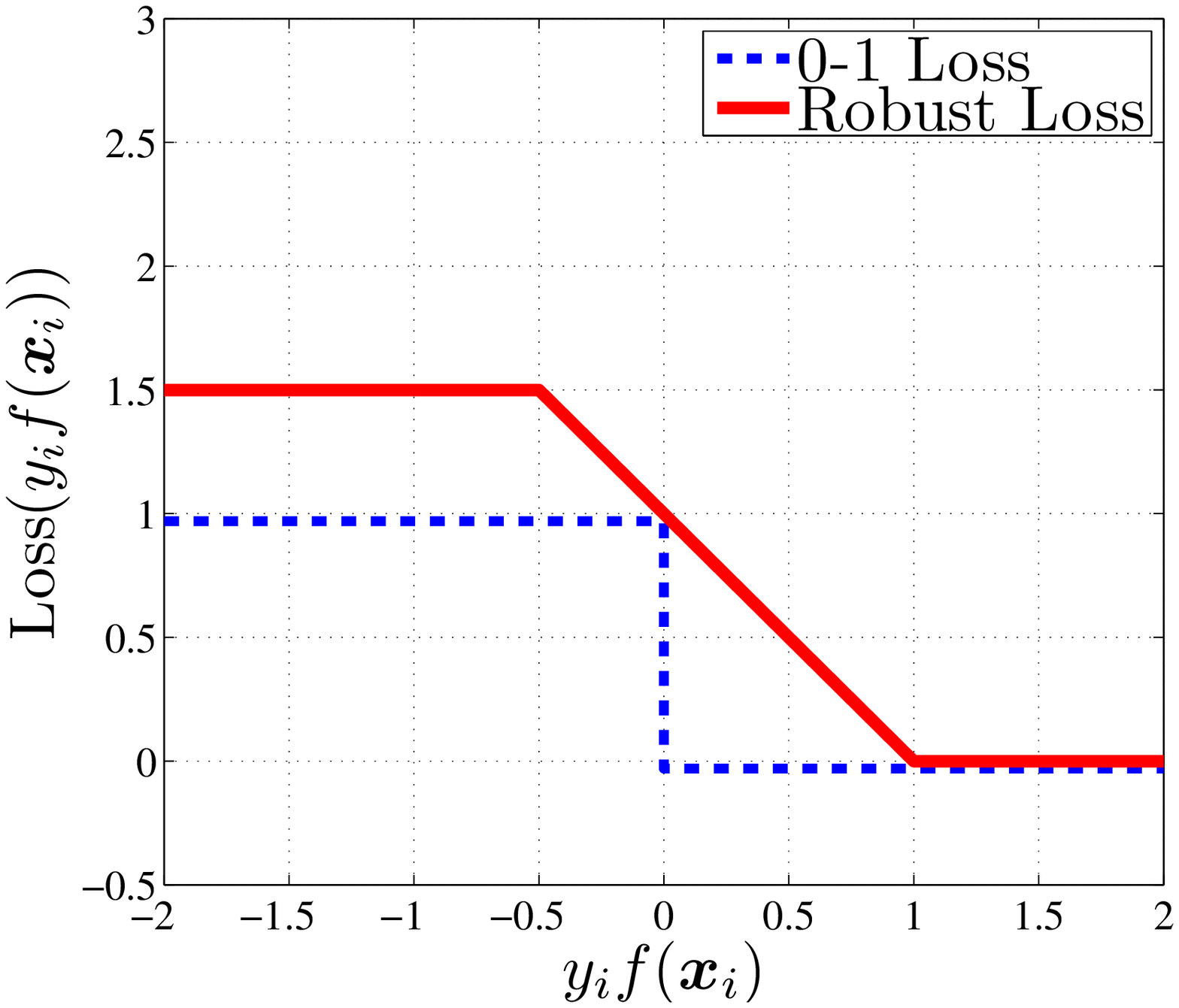} &
   \includegraphics[width=0.17\textwidth]{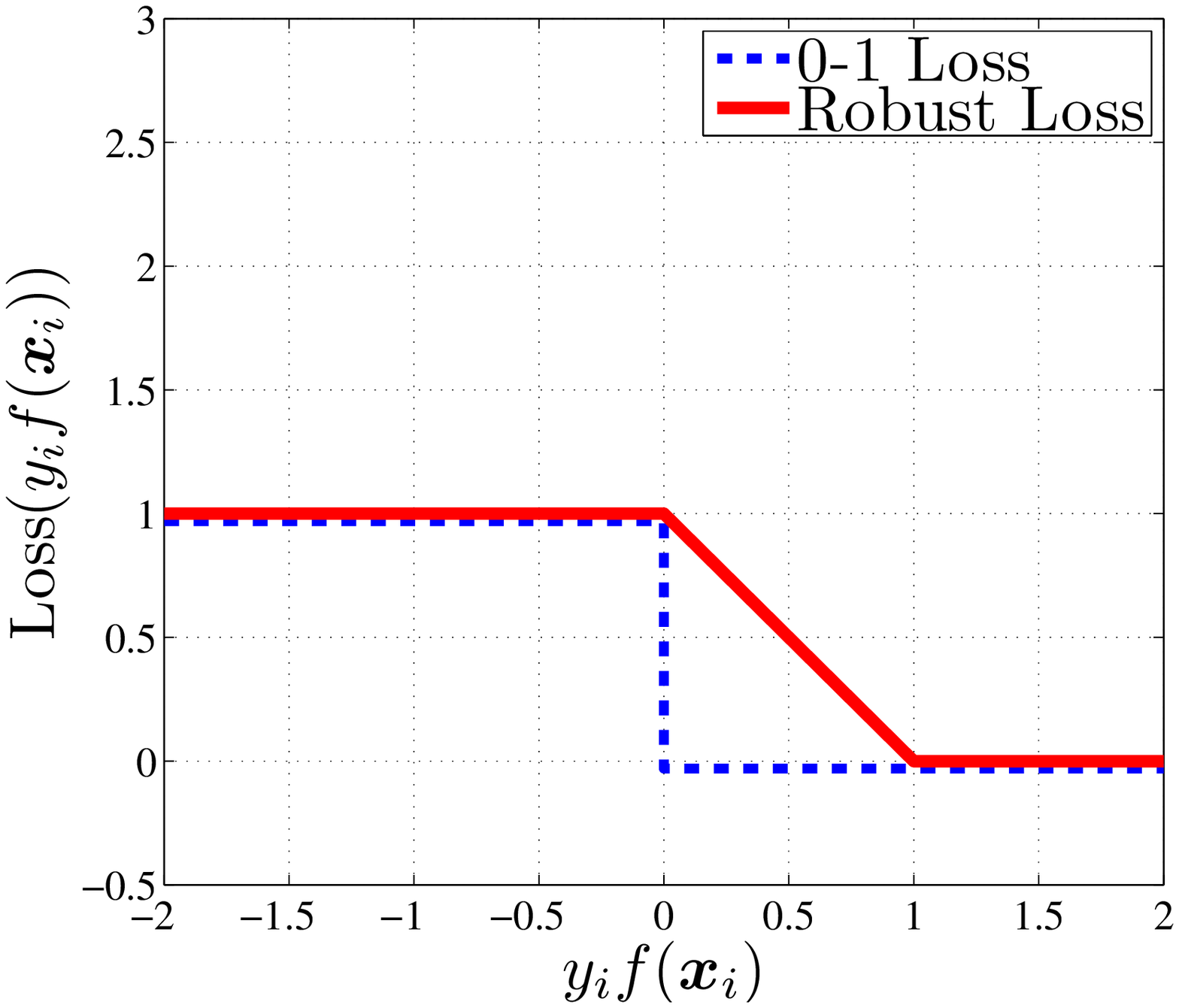} \\
   \footnotesize{$\theta = 0$, $s = -\infty$} &
   \footnotesize{$\theta = 0$, $s = -1.5$} &
   \footnotesize{$\theta = 0$, $s = -1$} &
   \footnotesize{$\theta = 0$, $s = -0.5$} &
   \footnotesize{$\theta = 0$, $s = 0$} \\		   
   \multicolumn{5}{c}{\small (b) Homotopy computation with decreasing $s$ from $-\infty$ to 0.} 
  \end{tabular}
  \caption{Robust loss functions for various homotopy parameters $\theta$ and $s$.}
  \label{fig:homotopy.illustration.svc}
  \begin{tabular}{ccccc}
   \includegraphics[width=0.17\textwidth]{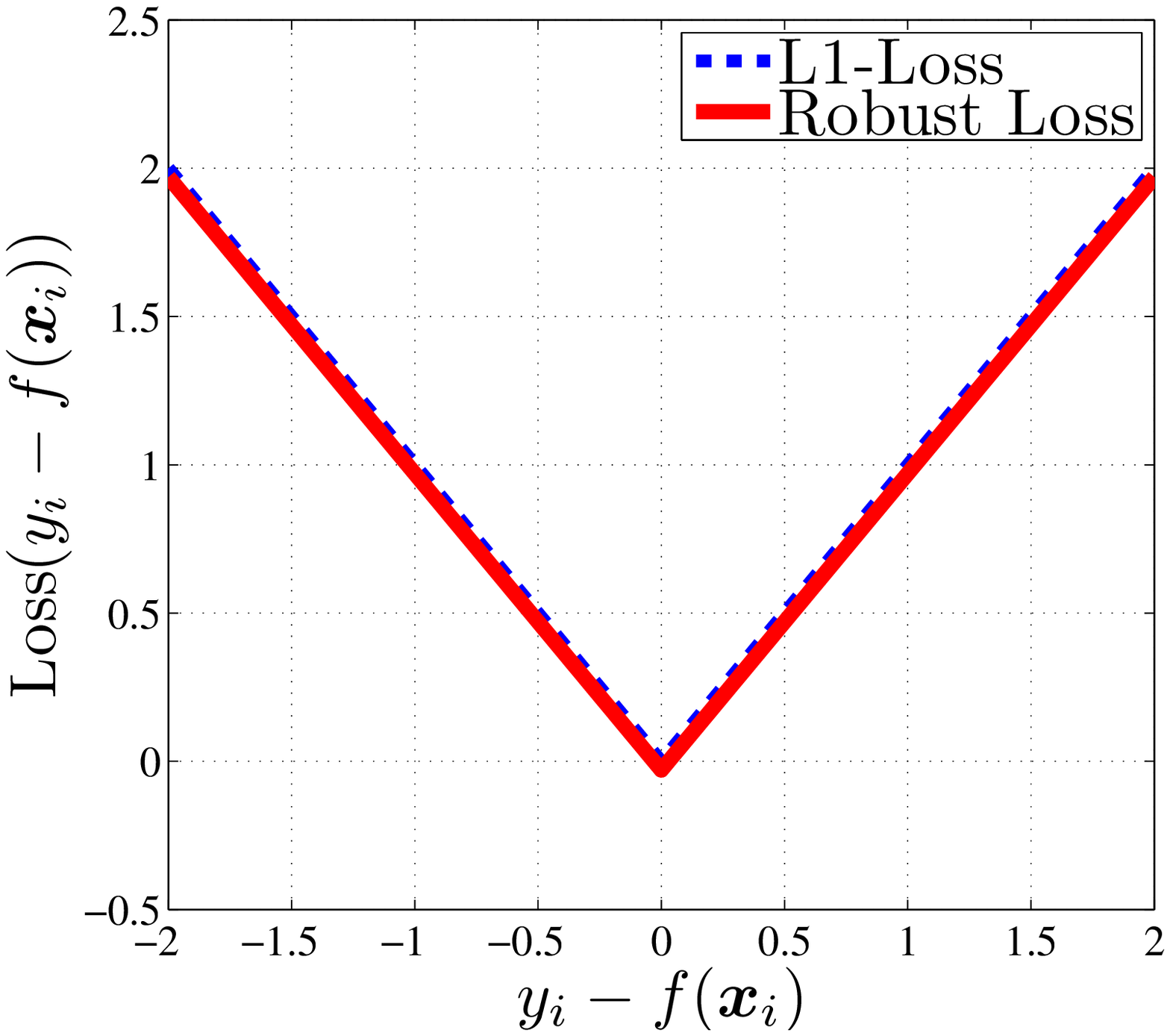} &
   \includegraphics[width=0.17\textwidth]{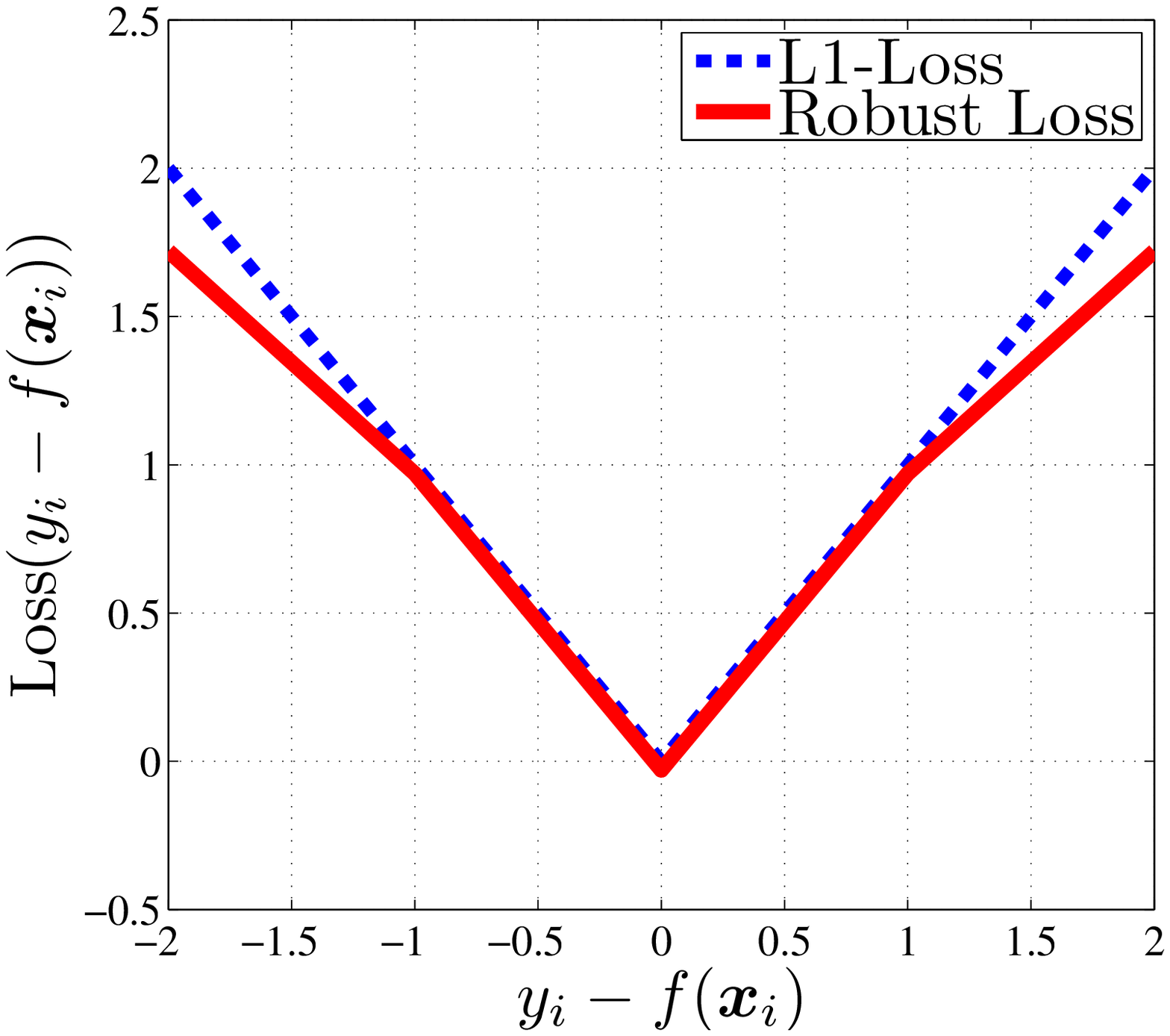} &
   \includegraphics[width=0.17\textwidth]{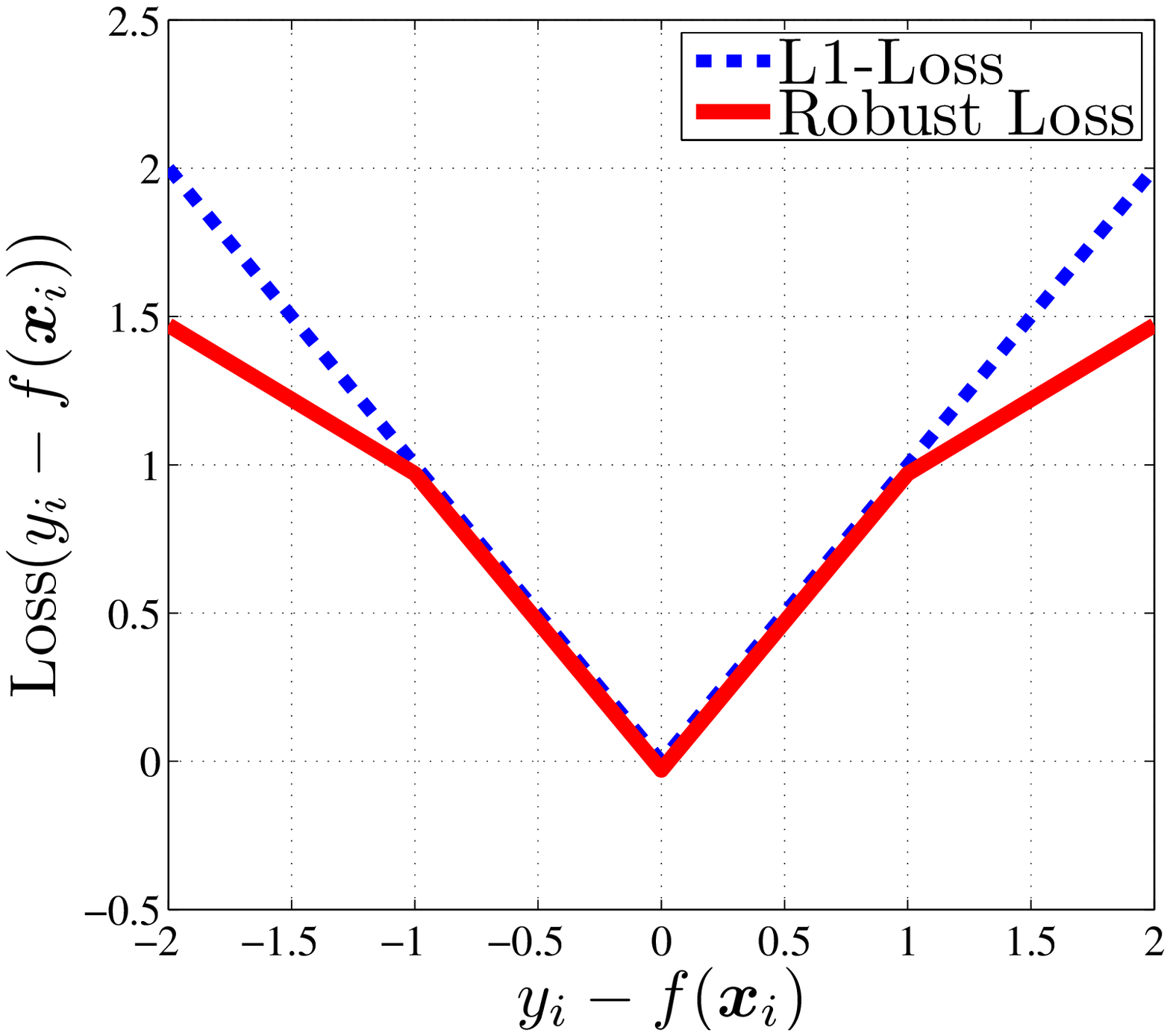} &
   \includegraphics[width=0.17\textwidth]{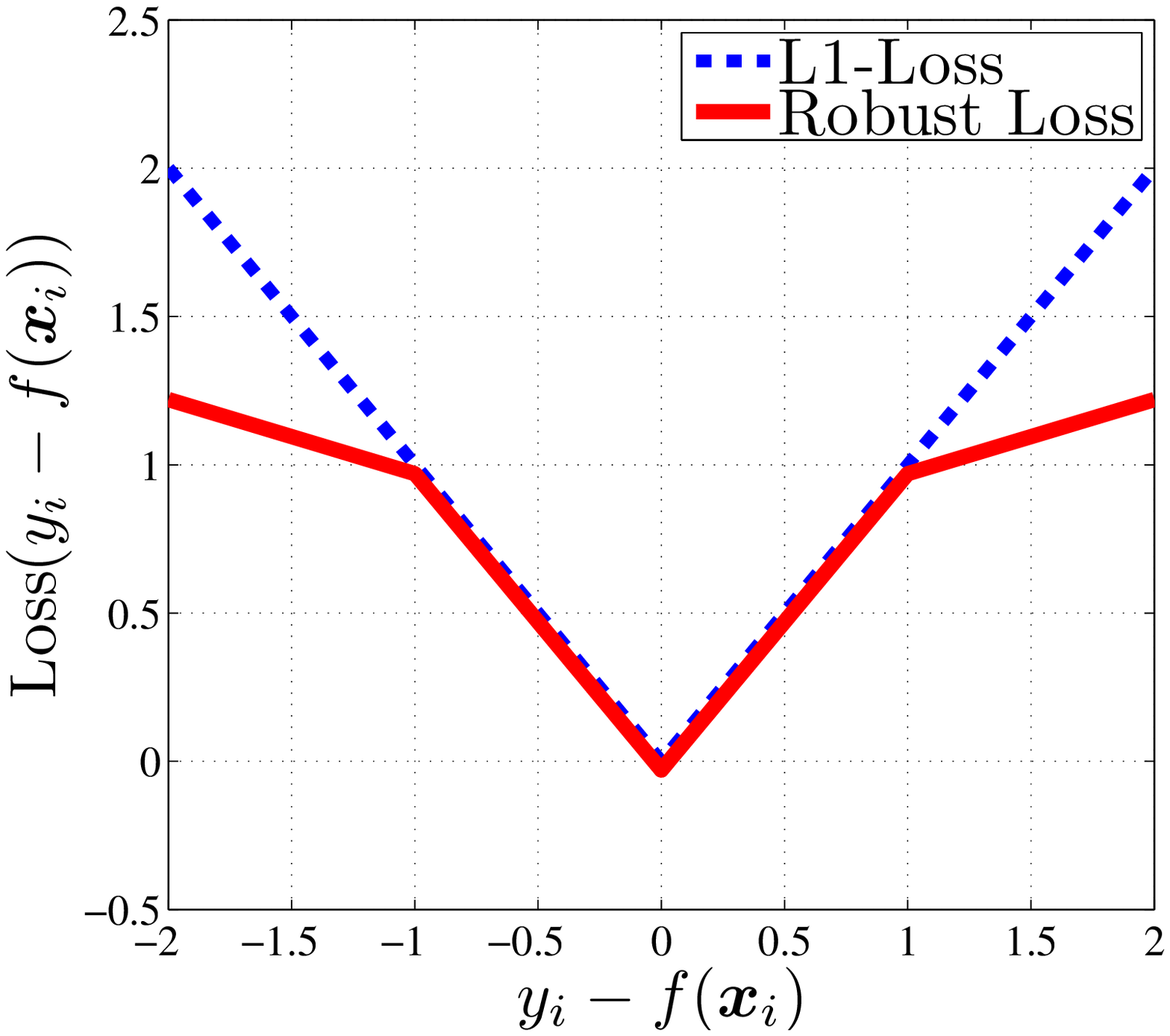} &
   \includegraphics[width=0.17\textwidth]{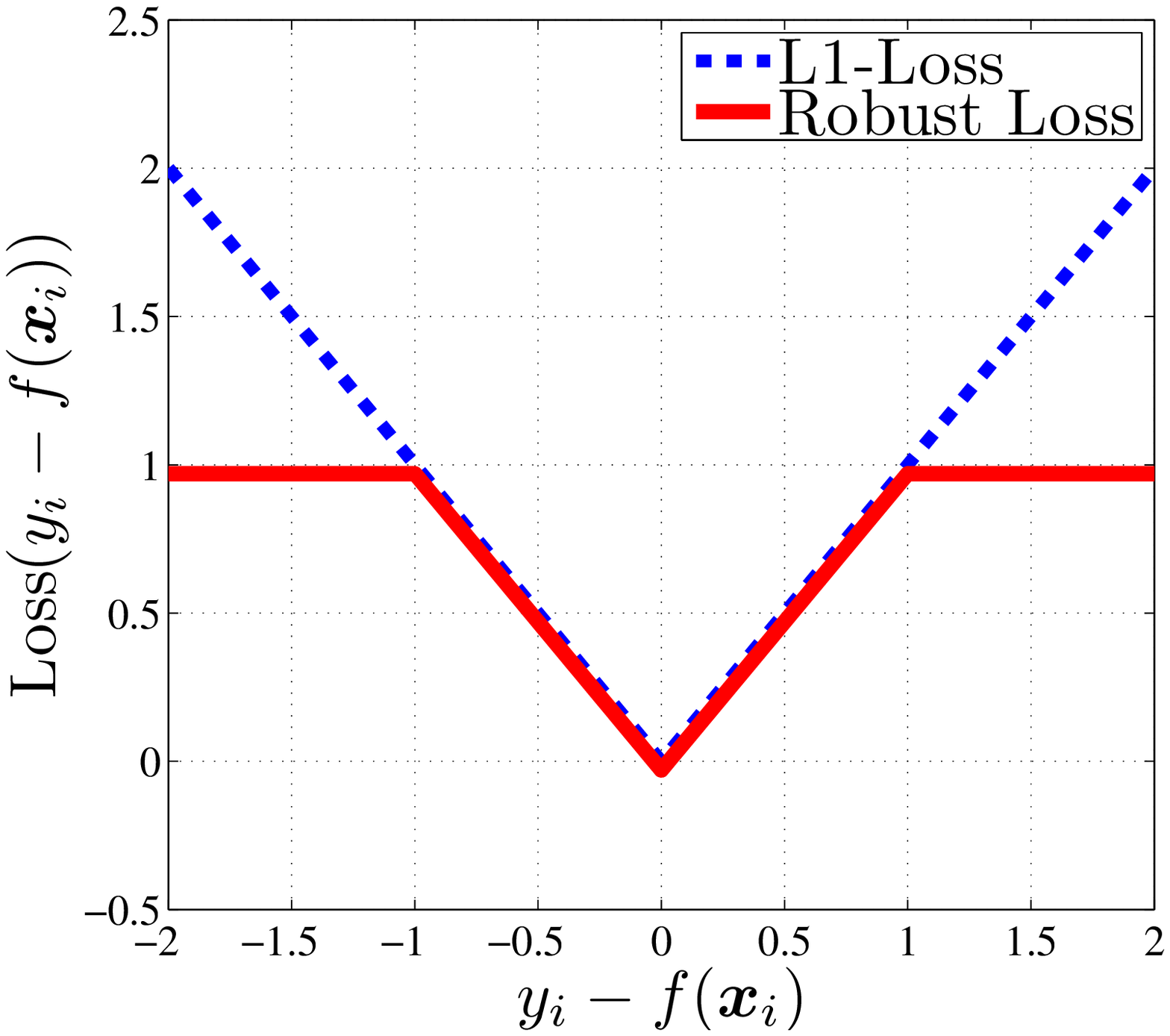} \\
   \footnotesize{$\theta = 1$, $s = 1$} &
   \footnotesize{$\theta = 0.75$, $s = 1$} &
   \footnotesize{$\theta = 0.5$, $s = 1$} &
   \footnotesize{$\theta = 0.25$, $s = 1$} &
   \footnotesize{$\theta = 0$, $s = 1$} \\		   
   \multicolumn{5}{c}{\small (a) Homotopy computation with decreasing $\theta$ from 1 to 0.} \\
   \includegraphics[width=0.17\textwidth]{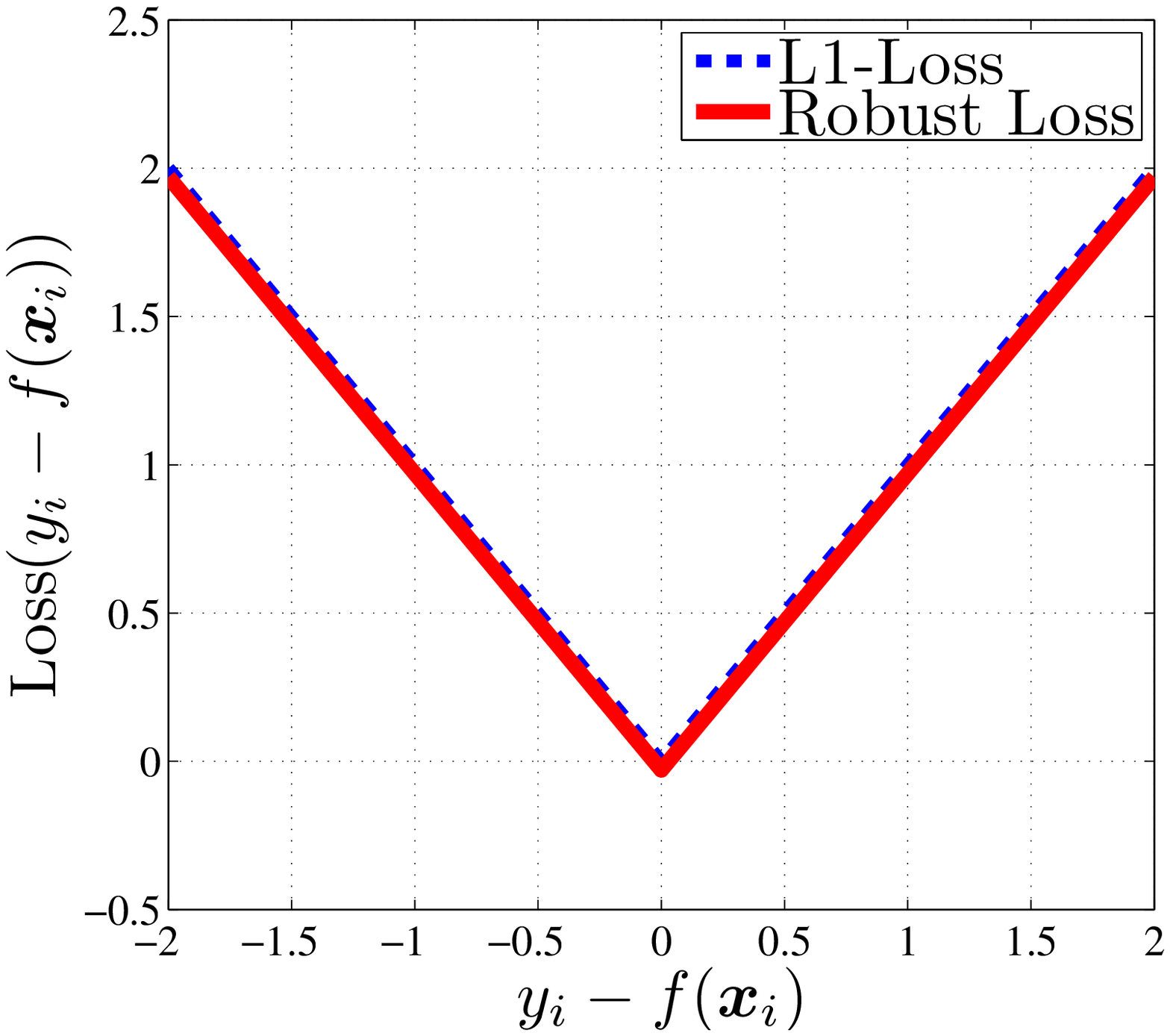} &
   \includegraphics[width=0.17\textwidth]{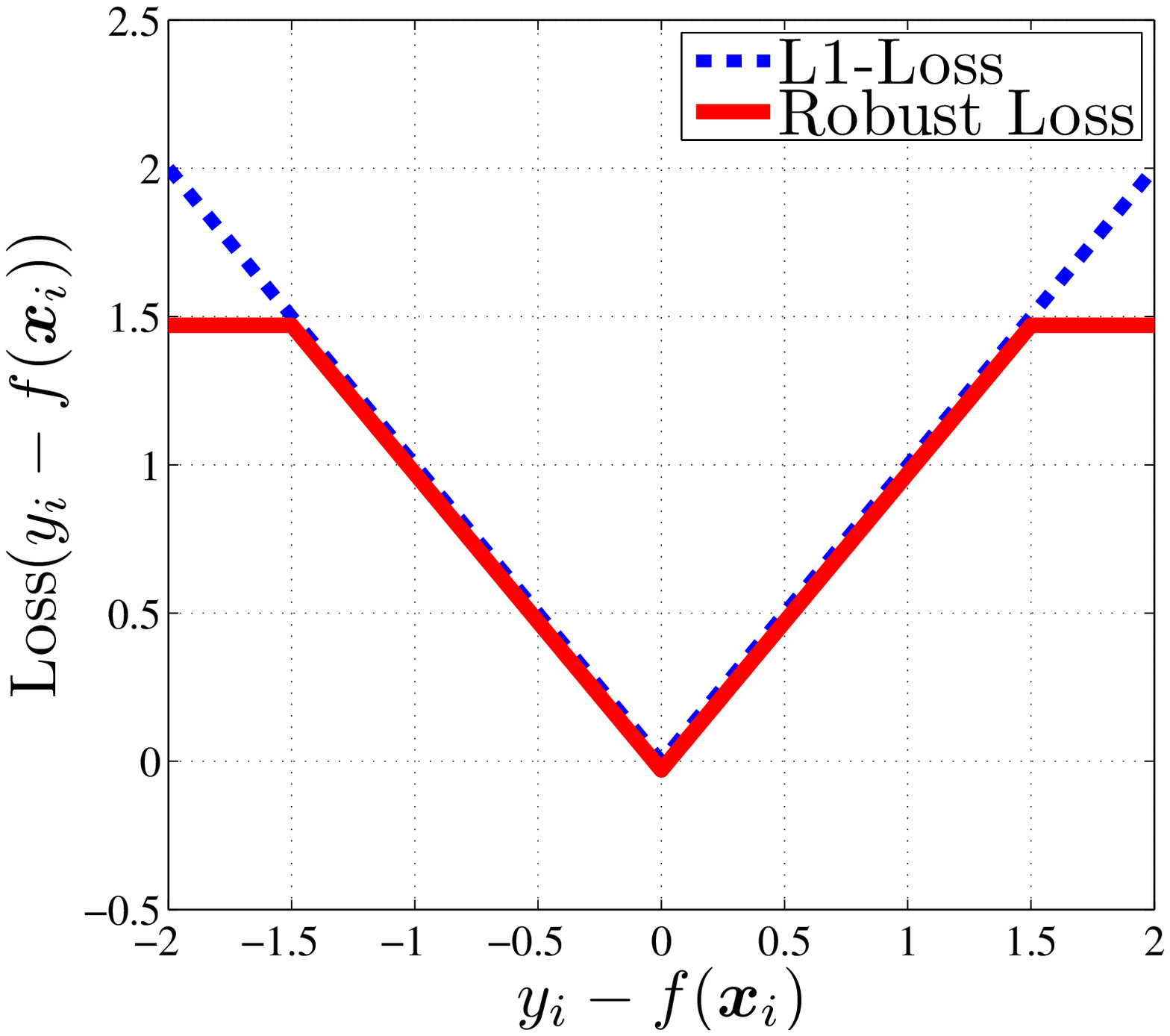} &
   \includegraphics[width=0.17\textwidth]{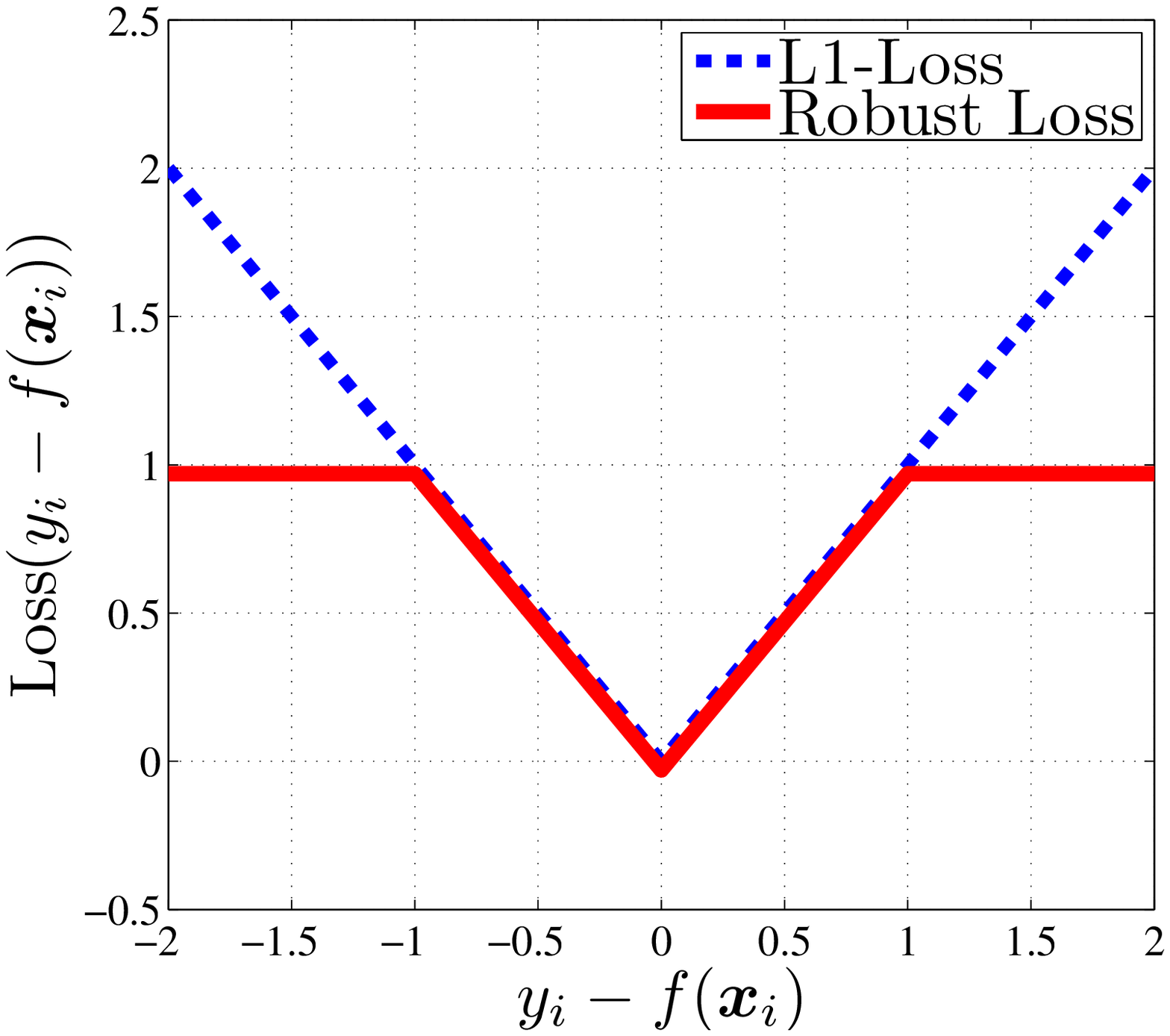} &
   \includegraphics[width=0.17\textwidth]{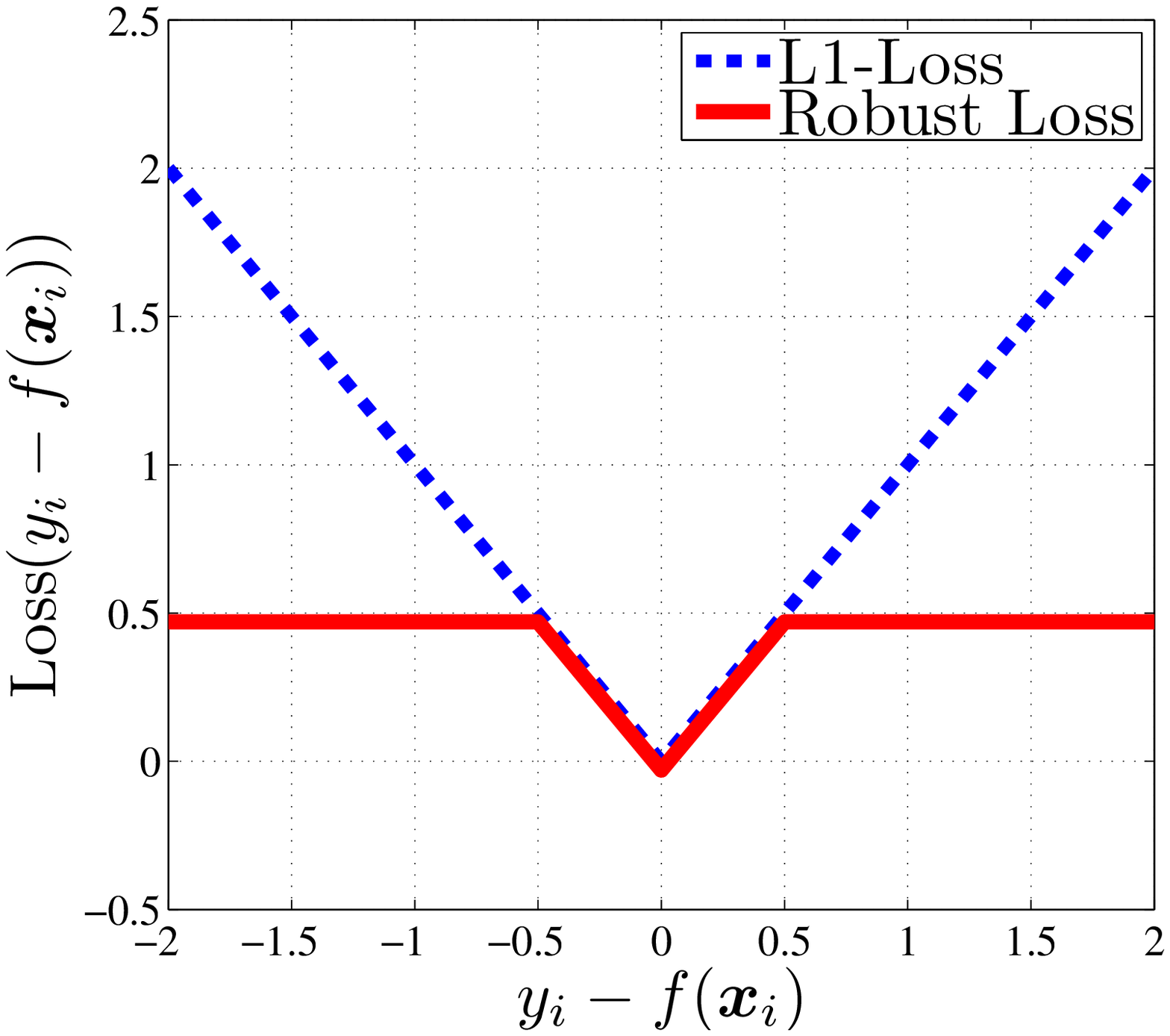} &
   \includegraphics[width=0.17\textwidth]{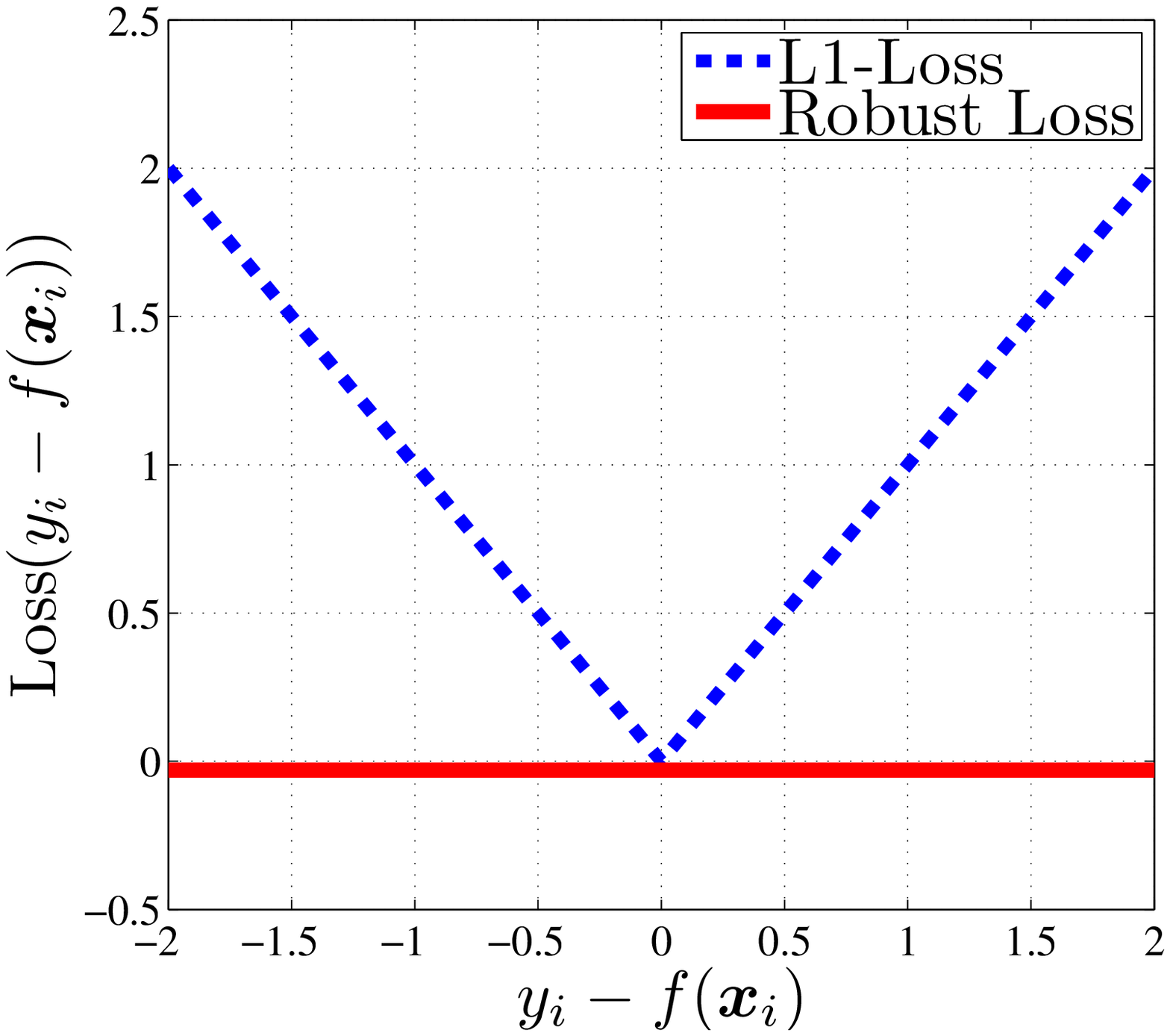} \\
   \footnotesize{$\theta = 0$, $s = \infty$} &
   \footnotesize{$\theta = 0$, $s = 1.5$} &
   \footnotesize{$\theta = 0$, $s = 1$} &
   \footnotesize{$\theta = 0$, $s = 0.5$} &
   \footnotesize{$\theta = 0$, $s = 0$} \\		   
   \multicolumn{5}{c}{\small (b) Homotopy computation with decreasing $s$ from $-\infty$ to 0.} 
  \end{tabular}
  \caption{Regression version of robust loss functions for various homotopy parameters $\theta$ and $s$.}
  \label{fig:homotopy.illustration.svr}
 \end{center}
\end{figure*}

First,
robust learning algorithms are essentially formulated 
as non-convex optimization problems
because the loss function must be designed to alleviate the effect of outliers
(see the robust loss functions for classification and regression problems in
\figurename \ref{fig:homotopy.illustration.svc}
and
\figurename
\ref{fig:homotopy.illustration.svr}). 
Since there could be many local optimal solutions 
in non-convex optimization problems,
it is important to develop non-convex optimization tricks 
such as simulated annealing \cite{Hromkovic01} that enables us to find good local optimal solution. 

The second practiall issue is
how to make a balance between robustness and efficiency.
Any robust SVM formulations contain 
an additional hyper-parameter
for controling the trade-off between robustness and efficiency.
Since such a hyper-parameter gorverns the influence of outliers on the model, 
it must be carefully tuned 
based on the property of noise contained in the data set.
In practice,
we need to empirically select an appropriate value for the hyper-parameter, 
e.g.,
by 
cross-validation, 
because 
we usually do not have sufficient knowledge about the noise in advance.
%
Furthermore, 
due to the non-convexity,
robust SVM solutions with slightly different hyper-parameter values can be significantly different,
which makes model selection highly unstable. 

In this paper,
we address these two issues simultaneously 
by introducing a novel \emph{homotopy} approach to
robust SV classification (SVC) and SV regression (SVR) learnings
\footnote{
For regression problems,
we study least absolute deviation (LAD) regression.
It is straightforward to extend it to original SVR formulation
with 
$\veps$-insensitive loss function. 
In order to simplify the description, 
we often call LAD regression as SV regression (SVR). 
In what follows, 
we use the term SVM 
when we describe common properties of SVC and SVR.
}.
Our basic idea is to consider parametrized formulations of robust SVC and SVR
which 
bridge 
the standard SVM
and
fully robust SVM
via a parameter 
that gorverns the influence of outliers.
We use
\emph{homotopy} methods
\cite{Allgower93,Gal95,Ritter84,Best96}
for tracing a path of solutions
when the influence of outliers is gradually decreased.
We call the parameter as the
\emph{robustness parameter}
and 
the path of solutions obtained by tracing the robustness parameter as the
\emph{robustification path}. 
\figurename
\ref{fig:homotopy.illustration.svc}
and 
\figurename
\ref{fig:homotopy.illustration.svr}
illustrate how the robust loss functions for classification and regression problems
can be gradually robustified,
respectively.

\subsection{Our contributions}
Our first technical contribution is
in analyzing the properties of the robustification path
for both classification and regression problems.
In particular, 
we derive the
necessary
and
sufficient
conditions for SVC and SVR solutions to be locally optimal
(note that the well-known Karush-Khun Tucker (KKT) conditions are only necessary, but not sufficient). 
Interestingly, 
the analyses indicate that 
the robustification paths contain a finite number of dicontinuous points. 
To the best of our knowledge, 
the above property of robust learning has not been known previously.

Our second contribution is to develop an efficient algorithm
for actually computing the robustification path
based on the above theoretical investigation of the geometry of robust SVM solutions.
Here,
we use parametric programming technique \cite{Allgower93,Gal95,Ritter84,Best96},
which is often used for computing the
\emph{regularization path} 
in machine learning literature.
The main technical challenge here is how to handle
the discontinuous points in the robustification path.
We develop an algorithm 
that can precisely detect such discontinuous points,
and \emph{jump} to find a strictly better local optimal solution.
Unlike solution path of convex problems \cite{Allgower93,Gal95,Ritter84,Best96},
our local optimal solution paths
is shown to have a finite number of discrete points.
We overcome this difficulty by precisely analyzing the necessary and
sufficient conditions for the local optimality.

Many existing studies on robust SVM employ
Concave Convex Procedure (CCCP) \cite{yuille02a}
or a variant called Difference of Convex (DC) programming \cite{Shen03,krause04a,Liu05a,Liu06a,Collobert06a,wu07a}. 
In these methods,
a non-convex loss function
is decomposed into the concave and convex parts.
A local optimal solution will be obtained by 
iteratively solving a sequence of convex optimization problems.
Local optimal solutions found by CCCP are sensitive to hyper-parameter values
(for controling the robustness and efficiency balance), 
which makes model selection highly challenging.

Experimental results indicate that
our outlier approach can find better
robust SVM solutions more efficiently
than alternative approaches based on CCCP. 
We conjecture that there are two reasons why favorable results can be obtained.
At first, 
the robustification path shares similar advantage to
\emph{simulated annealing} \cite{Hromkovic01}.
Simulated annealing is known to find better local solutions 
in many non-convex optimization problems 
by solving a sequence of solutions
along with 
so-called
\emph{temprature}
parameter. 
If we regard the robustness parameter as the temprature parameter, 
our robustification path algorithm can be interpreted as simulated annealing with infinitesimal step size. 
Another possible explanation for favorable performances of our method is
the ability of stable and efficient model selection. 
Since our algorithm provides the path of local solutions, 
unlike other non-convex optimization algorithms such as CCCP, 
two solutions with slightly different robustness parameter values tend to be similar,
which makes model selection stable. 
Accoding to our experiments,
choice of the robustness parameter is quite sensitive to the generalization performances. 
Thus,
it is important to finely tune the robustness parameter.
Since our algorithm can compute the path of solutions,
it is much more computationally efficient than running CCCP many times at different
robustness parameter values.

\subsection{Structure of This Paper}
After we formulate robust SVC and SVR as parametrized optimization problems
in 
\S~\ref{sec:Parameterized_Formulation_of_RSVM},
we derive in \S~\ref{sec:local-optimality}
the \emph{necessary} and \emph{sufficient} 
conditions for a robust SVM solution to be locally optimal,
and show that there exist a finite number of discontinuous points 
in the local solution path.
We then propose an efficient algorithm in \S~\ref{sec:algorithm}
that can precisely detect such discontinuous points
and \emph{jump} to find a strictly better local optimal solution. 
In \S~\ref{sec:experiment},
we experimentally demonstrate that our proposed method,
named the \emph{robustification path algorithm},
outperforms
the existing robust SVM algorithm based on CCCP
or DC programming.
Finally, we conclude in \S~\ref{sec:conclusions}.

This paper is an extented version of our preliminary conference paper presented at ICML 2014 \cite{suzumura2014outlier}. 
In this paper,
we have extended our robusitification path framework to the regression problem,
and many more experimental evaluations have been conducted. 
To the best of our knowledge,
the homotopy method \cite{Allgower93,Gal95,Ritter84,Best96} is first used in our preliminary conference paper
in the context of robust learning, 
So far,
homotopy-like methods have been (often implicitly) used for non-convex optimization problems 
in the context of sparse modeling \cite{zhang10a,mazumder11a,zhou12a}
and semi-supervised learning \cite{Ogawa13b}.

\section{Parameterized Formulation of Robust SVM}
\label{sec:Parameterized_Formulation_of_RSVM}
In this section,
we first formulate robust SVMs for classification and regression problems,
which we denote by
robust SVC (SV classification)
and
robust SVR (SV regression),
respectively.
Then,
we introduce parameterized formulation both for robust SVC and SVR,
where
the parameter gorverns the influence of outlieres to the model.
The problem is reduced to ordinary non-robust SVM 
at one end of the parameter, 
while 
the problem corresponds to fully-robust SVM 
at the other end of the parameter.
In the following sections,
we develop algorithms for computing the path of local optimal solutions 
when the parameter is changed form one end to the other. 

\subsection{Robust SV Classification}
\label{subsec:robust-SV-classification}
Let us consider a binary classification problem with $n$ instances and $d$ features. 
We denote the training set as 
$\{(\Vec{x}_i, y_i)\}_{i \in \NN_n}$
where
$\bm x_i \in \cX$
is the input vector in the input space $\cX \subset \RR^d$,
$y_i \in \{-1, 1\}$
is the binary class label,
and 
the notation 
$\NN_n := \{1, \ldots, n\}$
represents the set of natural numbers up to $n$.
We write the decision function as
\begin{align}
 \label{eq:decision-function}
 f(\Vec{x}) := \Vec{w}^\top \Vec{\phi}(\Vec{x}),
\end{align}
where \Vec{\phi} is the feature map implicitly defined by a kernel \Vec{K},
\Vec{w} is a vector in the feature space,
and $^\top$ denotes the transpose of vectors and matrices.

We introduce the following class of optimization problems
\emph{parameterized}
by
$\theta$
and
$s$:
\begin{align}
 \label{eq:robustSVM_training}
 \min_{\Vec{w}} ~ \frac{1}{2}\| \Vec{w} \|^2 + C \sum_{i = 1}^n \ell(y_i f(\Vec{x}_i); ~\theta, s),
\end{align}
where
$C > 0$
is the regularization parameter
which controls the balance
between the first regularization term and the second loss term.
The loss function
$\ell$
is characterized by a pair of parameters
$\theta \in [0, 1]$
and
$s \le 0$ as
\begin{align}
 \ell(z; ~\theta,s) := \left\{
	\begin{array}{cc}
	[0, 1 - z]_+, & z \ge s, \\
	1 - \theta z - s, & z < s, \\
	\end{array}
	\right.
\end{align}
where 
$[z]_+ := \max\{0, z\}$. 
We refer to
$\theta$
and
$s$
as
the \emph{homotopy parameters}. 
\figurename~\ref{fig:homotopy.illustration.svc}
shows the loss functions for several $\theta$ and $s$. 
The first homotopy parameter 
$\theta$
can be interpreted as the
\emph{weight}
for an outlier: 
$\theta = 1$
indicates that the influence of an outlier is the same as an inlier,
while 
$\theta = 0$
indicates that outliers are completely ignored.
The second homotopy parameter
$s \le 0$
can be interpreted as the threshold for deciding outliers and inliers.

In the following sections,
we consider two types of homotopy methods.
In the first method,
we fix $s = 0$,
and gradually change
$\theta$
from
1 to 0
(see the top five plots in \figurename~\ref{fig:homotopy.illustration.svc}). 
In the second method,
we fix
$\theta = 0$
and gradually change
$s$
from
$-\infty$
to 0
(see the bottom five plots in \figurename~\ref{fig:homotopy.illustration.svc}). 
Note that 
the loss function is reduced to 
the hinge loss for the standard (convex) SVC
when 
$\theta = 1$
or
$s = - \infty$.
Therefore,
each of the above two homotopy methods can be interpreted
as the process of tracing a sequence of solutions when 
the optimization problem is gradually modified from convex to non-convex. 
By doing so, we expect to find good local optimal solutions because such a process can be interpreted as
\emph{simulated annealing} ~\cite{Hromkovic01}.
In addition,
we can adaptively control the degree of robustness
by selecting the best
$\theta$
or 
$s$
based on some model selection scheme. 

\subsection{Robust SV Regression}
\label{subsec:robust-sv-regression}
Let us next consider a regression problem. 
We denote the training set of the regression problem as
$\{(\bm x_i, y_i)\}_{i \in \NN_n}$,
where the input
$\bm x_i \in \cX$
is the input vector as the classification case,
while
the output
$y_i \in \RR$
is a real scalar.
We consider a regression function
$f(\bm x)$
in the form of
\eq{eq:decision-function}.
SV regression is formulated as 
\begin{align}
 \label{eq:robustSVR_training}
 \min_{\Vec{w}}
 ~
 \frac{1}{2}\| \Vec{w} \|^2 + C \sum_{i = 1}^n \ell(y_i - f(\Vec{x}_i); ~\theta, s),
\end{align}
where
$C > 0$
is the regularization parameter,
and the loss function $\ell$ is defined as
\begin{align}
 \label{eq:regression-loss-func}
 \ell(z; ~\theta,s) := \left\{
	\begin{array}{ll}
	|z|, & |z| < s, \\
	(|z| - s) \theta + s, & |z| \ge s. \\
	\end{array}
	\right.
\end{align}
The loss function
in
\eq{eq:regression-loss-func}
has two parameters
$\theta \in [0, 1]$
and
$s \in [0, \infty)$
as the classification case.
\figurename~\ref{fig:homotopy.illustration.svr}
shows the loss functions for several $\theta$ and $s$.

\section{Local Optimality}
\label{sec:local-optimality}
In order to use the homotopy approach,
we need to clarify the continuity of the local solution path.
To this end,
we investigate several properties of local solutions of robust SVM,
and derive the necessary and sufficient conditions.
Interestingly, 
our analysis reveals that the local solution path has a finite number of
\emph{discontinuous}
points. 
The theoretical results presented here form the basis of
our novel homotopy algorithm given in the next section 
that can properly handle the above discontinuity issue.
We first
discuss the local optimality of robust SVC
in detail
in
\S~\ref{subsec:cond-opt-sol}
and
\S~\ref{subsec:n-and-s-conditions},
and then present the corresponding result of robust SVR briefly in
\S~\ref{subsec:local-opt-regression}.

\subsection{Conditionally Optimal Solutions (for Robust SVC)}
\label{subsec:cond-opt-sol}
The basic idea of our theoretical analysis is
to reformulate the robust SVC learning problem
as a combinatorial optimization problem.
We consider a partition of the instances
$\NN_n := \{1, \ldots, n\}$
into two disjoint sets 
$\cI$
and
$\cO$.
The instances in
$\cI$
and
$\cO$
are defined as 
$\cI$nliers
and 
$\cO$utliers,
respectively.
Here, 
we restrict that the margin
$y_i f(\bm x_i)$
of an inlier should be larger than $s$,
while
that of an outlier should be smaller than $s$.
We denote the partition as
$\cP := \{\cI, \cO\} \in 2^{\NN_n}$,
where
$2^{\NN_n}$
is the power set\footnote{
The power set means that
there are $2^n$ patterns that each of the instances belongs to either $\cI$ or $\cO$.
} of $\NN_n$. 
Given a partition 
$\cP$, 
the above restrictions define the feasible region of the solution $f$ 
in the form of a convex polytope\footnote{Note that an instance with the margin $y_i f(\bm{x}_i) = s$ can be the member of either $\cI$ or $\cO$.}:
\begin{align}
 \label{eq:polytope}
 {\rm pol}({\cP};s) := \Biggl\{f 
 ~ \Biggl| 
 \begin{array}{ll}
  y_i f(\Vec{x}_i) \ge s, & i \in \cI \\
  y_i f(\Vec{x}_i) \le s, & i \in \cO
 \end{array}
 \Biggr\}.
\end{align}
Using the notion of the convex polytopes, 
the optimization problem
\eq{eq:robustSVM_training}
can be rewritten as
\begin{align}
 \label{eq:combinatorial.view.1}
 \min_{{\cP} \in 2^{\NN_n}}
 \Biggl( \min_{f \in {\rm pol}({\cP};s)} J_{{\cP}}(f;\theta) \Biggr),
\end{align}
where the objective function
$J_{{\cP}}$
is defined as\footnote{Note that we omitted the constant terms irrelevant to the optimization problem.}
\begin{align*}
J_{{\cP}}(f;\theta)
   := \frac{1}{2} || \Vec{w} ||_2^2
  + C \left(
  \sum_{i \in \cI}[1 - y_i f(\Vec{x}_i)]_+ +
  \theta \sum_{i \in \cO}[1 - y_i f(\Vec{x}_i)]_+
\right).
\end{align*}

When the partition
${\cP}$
is fixed,
it is easy to confirm that the inner minimization problem of
\eq{eq:combinatorial.view.1}
is a convex problem.

\vspace*{12pt}
\begin{defi}[Conditionally optimal solutions]
 Given a partition
 ${\cP}$, 
 the solution of the following convex problem is said to be
 the conditionally optimal solution:
 \begin{align}
  \label{eq:conditionally.optimal.solution}
  f^*_{{\cP}} := \argmin_{f \in {\rm pol}({\cP};s)} J_{{\cP}}(f;\theta).
 \end{align}
\end{defi}
\vspace*{12pt}
The formulation in
\eq{eq:combinatorial.view.1}
is interpreted as a combinatorial optimization problem of 
finding the best solution from all the
$2^n$
conditionally optimal solutions
$f^*_{{\cP}}$
corresponding to all possible
$2^n$
partitions\footnote{
For some partitions
${\cP}$,
the convex problem
\eq{eq:conditionally.optimal.solution}
might not have any feasible solutions. 
}.

Using the representer theorem or convex optimization theory, 
we can show that any conditionally optimal solution can be written as
\begin{align}
 f^*_{{\cP}}(\bm x) := \sum_{j \in \NN_n} \alpha^*_j y_j K(\bm x, \bm x_j),
\end{align}
where 
$\{\alpha^*_j\}_{j \in \NN_n}$
are the optimal Lagrange multipliers. 
The following lemma summarizes the KKT optimality conditions of
the conditionally optimal solution $f^*_{\cP}$. 

\vspace*{12pt}
\begin{lemm}
The KKT conditions of the convex problem 
\eq{eq:conditionally.optimal.solution}
is written as 
\begin{subequations}
 \label{eq:KKT_conds}
\begin{align}
 y_i f^*_{{\cP}}(\Vec{x}_i) > 1 ~ & \Rightarrow ~ \alpha_i^* = 0, \\
 y_i f^*_{{\cP}}(\Vec{x}_i) = 1 ~ & \Rightarrow ~ \alpha_i^* \in [0, C], \\
 s < y_i f^*_{{\cP}}(\Vec{x}_i) < 1 ~ & \Rightarrow ~ \alpha_i^* = C, \\
 y_i f^*_{{\cP}}(\Vec{x}_i) = s, i \in \cI ~ & \Rightarrow ~ \alpha_i^* \ge C, \\
 y_i f^*_{{\cP}}(\Vec{x}_i) = s, i \in \cO ~ & \Rightarrow ~ \alpha_i^* \le C \theta, \\
 y_i f^*_{{\cP}}(\Vec{x}_i) < s ~ & \Rightarrow ~ \alpha_i^* = C \theta.
\end{align}
\end{subequations}
\end{lemm}
\vspace*{12pt}
The proof is omitted because it can be easily derived based on standard convex optimization theory
\cite{Boyd04a}.

\subsection{The necessary and sufficient conditions for local optimality (for Robust SVC)}
\label{subsec:n-and-s-conditions}
From the definition of conditionally optimal solutions,
it is clear that 
a local optimal solution must be conditionally optimal within the convex polytope ${\rm pol}({\cP};s)$.
However, the conditional optimality does not necessarily indicate the local optimality
as the following theorem suggests. 
\vspace*{12pt}
\begin{theo}
 \label{theo:label.flip}
 For any
 $\theta \in [0, 1)$ and $s \le 0$,
 consider the situation where a conditionally optimal solution 
 $f^*_{{\cP}}$
 is at the boundary of the convex polytope
 ${\rm pol}({\cP};s)$,
 i.e.,
 there exists at least an instance such that
 $y_i f^*_{{\cP}}(\bm x_i) = s$. 
 In this situation,
 if we define a new partition
 $\tilde{\cP} := \{\tilde{\cI}, \tilde{\cO}\}$
 as 
\begin{subequations}
 \label{eq:new.I.O}
 \begin{eqnarray}
 \tilde{\cI} \!\leftarrow\! \cI \!\setminus\! \{ i \in \cI | y_i f^*(\Vec{x}_i) \!=\! s \} \!\cup\! \{ i \in \cO | y_i f^*(\Vec{x}_i) \!=\! s \}, \\
 \tilde{\cO} \!\leftarrow\! \cO \!\setminus\! \{ i \in \cO | y_i f^*(\Vec{x}_i) \!=\! s \} \!\cup\! \{ i \in \cI | y_i f^*(\Vec{x}_i) \!=\! s \},
 \end{eqnarray}
\end{subequations}
 then
 the new conditionally optimal solution 
 $f^*_{\tilde{\cP}}$
 is strictly better than
 the original conditionally optimal solution 
 $f^*_{\cP}$, 
 i.e.,
 \begin{align}
  \label{eq:strictly_better}
  J_{\tilde{\cP}}(f^*_{\tilde{\cP}};\theta) < J_{\cP}(f^*_{\cP};\theta).
 \end{align}
\end{theo}
\vspace*{12pt}
The proof is presented in Appendix \ref{app:lemm:strictly.better}.
Theorem~\ref{theo:label.flip}
indicates that
if 
$f^*_{{\cP}}$
is at the boundary of the convex polytope
${\rm pol}({\cP};s)$,
i.e., 
if there is one or more instances such that 
$y_i f^*_{\cP} (\bm x_i) = s$,
then 
$f^*_{{\cP}}$
is NOT locally optimal 
because there is a strictly better solution in the opposite side of the boundary. 

\begin{figure}[!h]
  \begin{tabular}{cc}
  \begin{minipage}{0.5\hsize}
    \begin{center}
    \subfigure[Local solution path] {
      \includegraphics[width=0.65\linewidth]{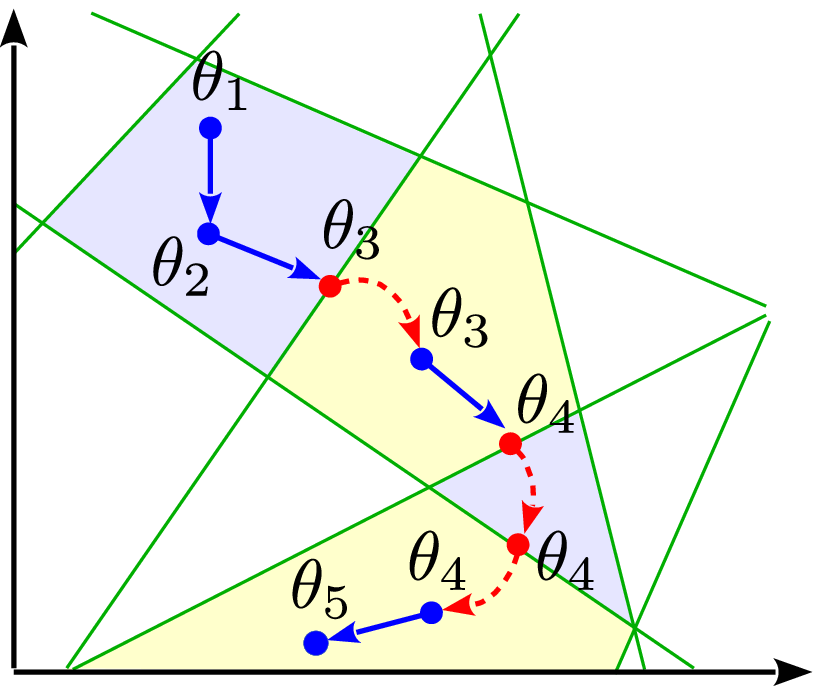}
     \label{fig:solution_a}
    }
    \end{center}
  \end{minipage}
  \begin{minipage}{0.5\hsize}
    \begin{center}
    \subfigure[Local optimum] {
      \includegraphics[width=0.65\linewidth]{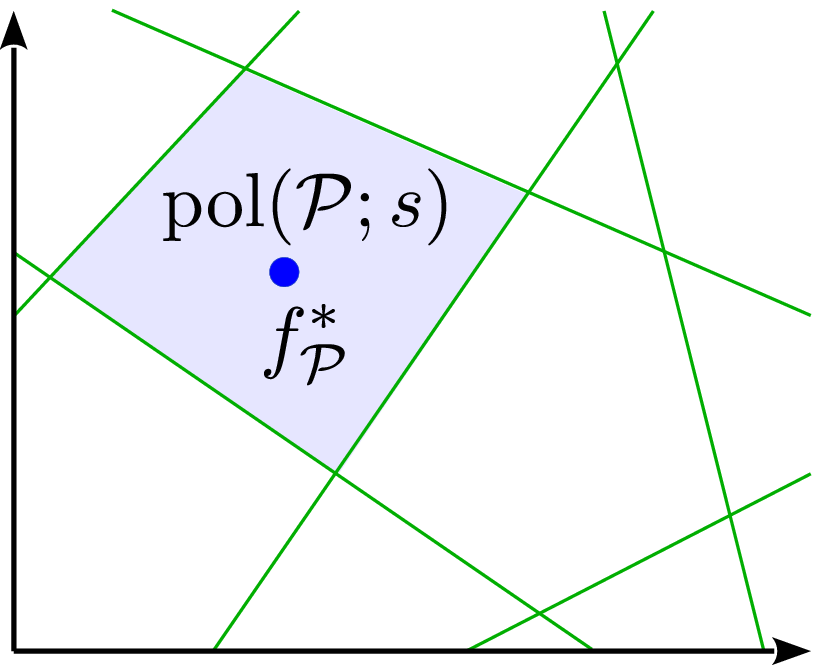}
      \label{fig:solution_b}
    }
    \end{center}
  \end{minipage}
  \end{tabular}
  \begin{tabular}{cc}
  \begin{minipage}{0.5\hsize}
    \begin{center}
    \subfigure[Not local optimum] {
      \includegraphics[width=0.65\linewidth]{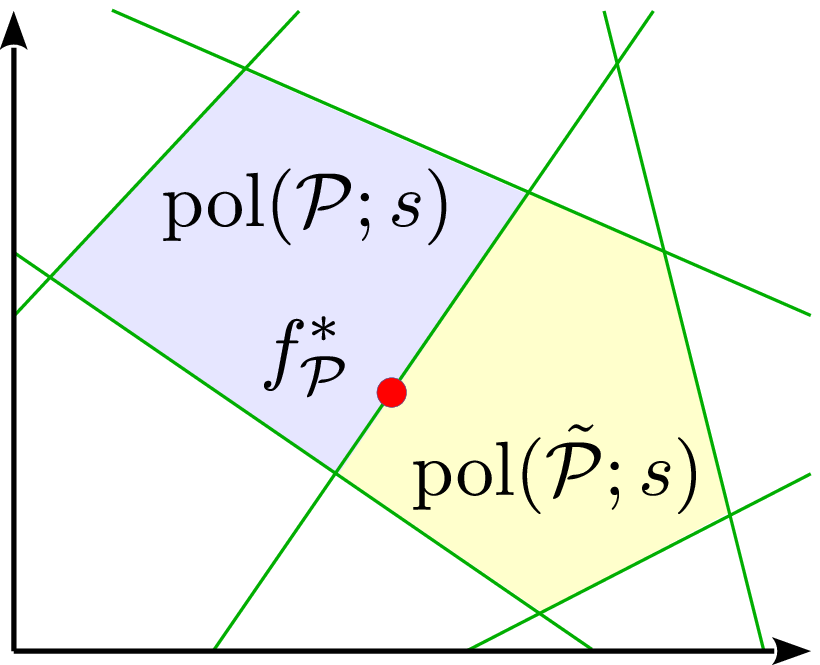}
      \label{fig:solution_c}
    }
    \end{center}
  \end{minipage}
  \begin{minipage}{0.5\hsize}
    \begin{center}
    \subfigure[Local optimum] {
      \includegraphics[width=0.65\linewidth]{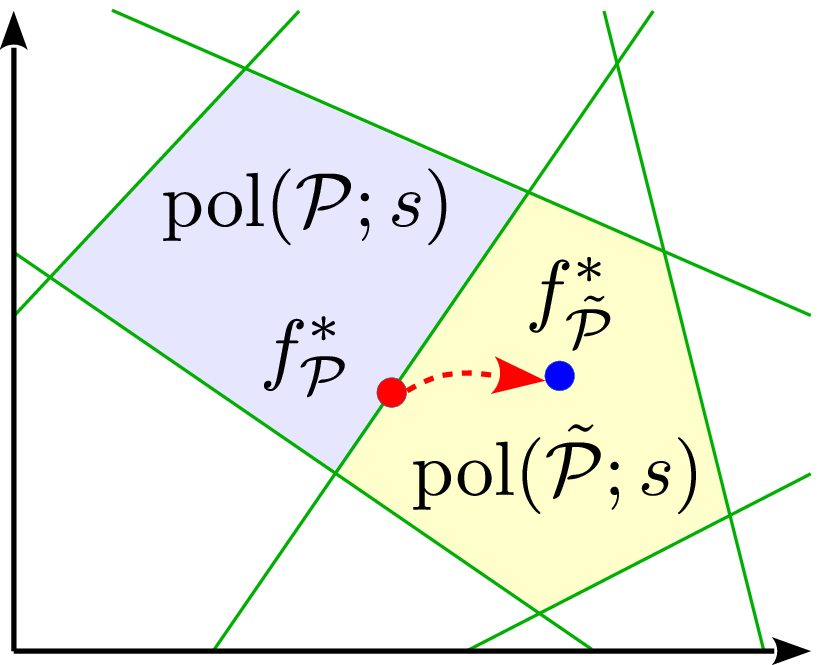}
      \label{fig:solution_d}
    }
    \end{center}
  \end{minipage}
  \end{tabular}
  \caption{Solution space of robust SVC.
 (a) The arrows indicate a local solution path when $\theta$ is gradually
    moved from $\theta_1$ to $\theta_5$ (see \S~\ref{sec:algorithm} for
    more details).
 (b) $f^*_\cP$ is locally optimal if it is at the strict interior of the convex polytope ${\rm pol}({\cP};s)$.
 (c) If $f^*_\cP$ exists at the boundary, then $f^*_\cP$ is feasible, but not locally optimal.
     A new convex polytope ${\rm pol}({\tilde{\cP}};s)$ defined in the
     opposite side of the boundary is shown in yellow.
 (d) A strictly better solution exists in ${\rm pol}({\tilde{\cP}};s)$.}
 \label{fig:solution_space}
\end{figure}

The following theorem summarizes the necessary and sufficient conditions
for local optimality.
Note that,
in non-convex optimization problems, 
the KKT conditions are necessary but not sufficient in general.
\vspace*{12pt}
\begin{theo}
 \label{theo:local.optimality.condition}
 For 
 $\theta \in [0, 1)$ and $s \le 0$,
 \begin{subequations}
  \label{eq:local.optimality.condition}
   \begin{eqnarray}
    \label{eq:local.optimality.condition.a}
     y_i f^*(\Vec{x}_i) > 1 &\; \Rightarrow \;& \alpha_i^* = 0, \\
    \label{eq:local.optimality.condition.b}
     y_i f^*(\Vec{x}_i) = 1 &\; \Rightarrow \;& \alpha_i^* \in [0, C], \\
    \label{eq:local.optimality.condition.c}
     s < y_i f^*(\Vec{x}_i) < 1 &\; \Rightarrow \;& \alpha_i^* = C, \\
    \label{eq:local.optimality.condition.d}
     y_i f^*(\Vec{x}_i) < s &\; \Rightarrow \;& \alpha_i^* = C \theta, \\
     \label{eq:local.optimality.condition.f}
     y_i f^*(\Vec{x}_i) \neq s, &\; \;& \!\!\!\!\! \!\!\!\!\! \!\!\!\!\! \forall i \in \NN_n,
   \end{eqnarray}
 \end{subequations}
 are necessary and sufficient for $f^*$ to be locally optimal.
\end{theo}
\vspace*{12pt}
The proof is presented in Appendix \ref{app:theo:local.optimality.condition}.
The condition
\eq{eq:local.optimality.condition.f}
indicates that the solution at the boundary of the convex polytope is not locally optimal. 
\figurename~\ref{fig:solution_space}
illustrates when a conditionally optimal solution can be locally optimal with a certain $\theta$ or $s$.

Theorem~\ref{theo:local.optimality.condition} suggests that, 
whenever the local solution path
computed by the homotopy approach encounters a boundary of the current convex polytope at a certain $\theta$ or $s$, 
the solution is not anymore locally optimal. 
In such cases,
we need to somehow find a new local optimal solution at that $\theta$ or $s$,
and restart the local solution path from the new one. 
In other words,
the local solution path has \emph{discontinuity} at that $\theta$ or $s$. 
Fortunately,
Theorem~\ref{theo:label.flip}
tells us how to handle such a situation. 
If the local solution path arrives at the boundary,
it can \emph{jump} to the new conditionally optimal solution
$f^*_{\tilde{\cP}}$ which is located on the opposite side of the boundary. 
This jump operation is justified because the new solution is shown to be strictly better than the previous one.
\figurename~\ref{fig:solution_space} (c) and (d)
illustrate such a situation.

\subsection{Local optimality of SV Regression}
\label{subsec:local-opt-regression}
In order to derive
the necessary and sufficent conditions
of the local optimality
in robust SVR, 
with abuse of notation, 
let us consider a partition of the instances
$\NN_n$
into two disjoint sets 
$\mathcal{I}$
and
$\mathcal{O}$,
which represent inliers and outliers,
respectively.
In regression problems,
an instance
$(\bm x_i, y_i)$
is regarded as an outlier
if the aboslute residual 
$|y_i - f(\bm x_i)|$
is sufficiently large.
Thus,
we define inliers and outliers of regression problem as 
\begin{eqnarray*}
  \mathcal{I} &:=& \{i \in \mathbb{N}_n |~ |y_i - f(\Vec{x}_i)| < s\}, \\
  \mathcal{O} &:=& \{i \in \mathbb{N}_n |~ |y_i - f(\Vec{x}_i)| > s\}.
\end{eqnarray*}
Given a partition
$\cP := \{\cI, \cO\} \in 2^{\NN_n}$,
the feasible region of the solution
$f$
is represented as a convex polytope:
\begin{align}
 {\rm pol}(\cP; s) := \Biggl\{f 
 ~ \Biggl| 
 \begin{array}{ll}
  |y_i - f(\Vec{x}_i)| \le s, & i \in \mathcal{I}, \\
  |y_i - f(\Vec{x}_i)| \ge s, & i \in \mathcal{O}
 \end{array}
 \Biggr\}.
\end{align}

Then,
as in the classification case, 
the optimization problem
\eq{eq:robustSVR_training}
can be rewritten as
\begin{align}
 \label{eq:combinatorial.view.1_svr}
 \min_{\cP}
 \Biggl( \min_{f \in {\rm pol}({\cP};s)} J_{{\cP}}(f;\theta) \Biggr),
\end{align}
where the objective function
$J_{\cP}$
is defined as
\begin{align*}
J_{\cP}(f; \theta)
  = \frac{1}{2} || \Vec{w} ||_2^2 + C \left(
  \sum_{i \in \mathcal{I}}|y_i - f(\Vec{x}_i)| +
  \theta \sum_{i \in \mathcal{O}}|y_i - f(\Vec{x}_i)|
\right).
\end{align*}
Since the inner problem of (\ref{eq:combinatorial.view.1_svr}) is a convex problem,
any conditionally optimal solution can be written as
\begin{align}
 f^*_{{\cP}}(\bm x) := \sum_{j \in \NN_n} \alpha^*_j K(\bm x, \bm x_j).
\end{align}
The KKT conditions of $f^*_{{\cP}}(\bm x)$ are written as
\begin{subequations}
\begin{align}
 |y_i - f^*_{{\cP}}(\Vec{x}_i)| = 0       ~ & \Rightarrow ~ 0 \le |\alpha_i^*| \le C, \\
 0 \le |y_i - f^*_{{\cP}}(\Vec{x}_i)| < s ~ & \Rightarrow ~ |\alpha_i^*| = C, \\
 |y_i - f^*_{{\cP}}(\Vec{x}_i)| = s, i \in \mathcal{I} ~ & \Rightarrow ~ |\alpha_i^*| \ge C, \\
 |y_i - f^*_{{\cP}}(\Vec{x}_i)| = s, i \in \mathcal{O} ~ & \Rightarrow ~ |\alpha_i^*| \le \theta C, \\
 |y_i - f^*_{{\cP}}(\Vec{x}_i)| > s ~ & \Rightarrow ~ |\alpha_i^*| = \theta C.
\end{align}
\end{subequations}

Based on the same discussion as
\S\ref{subsec:n-and-s-conditions}, 
the necessary and sufficient conditions for the local optimality of robust SVR
are summarized as the following theorem:
\begin{theo}
\begin{subequations}
 For 
 $\theta \in [0, 1)$ and $s \ge 0$,
\begin{align}
 |y_i - f^*_{{\cP}}(\Vec{x}_i)| = 0       ~ & \Rightarrow ~ 0 \le |\alpha_i^*| \le C, \\
 0 \le |y_i - f^*_{{\cP}}(\Vec{x}_i)| < s ~ & \Rightarrow ~ |\alpha_i^*| = C, \\
 |y_i - f^*_{{\cP}}(\Vec{x}_i)| > s       ~ & \Rightarrow ~ |\alpha_i^*| = \theta C, \\
\label{eq:local.optimality.condition.f.rsvr}
 |y_i - f^*_{{\cP}}(\Vec{x}_i)| \neq s.
\end{align}
are necessary and sufficient for $f^*$ to be locally optimal.
\end{subequations}
\end{theo}
We omit the proof of this theorem because they can be easily derived
in the same way as
Theorem~\ref{theo:local.optimality.condition}.

\section{Outlier Path Algorithm}
\label{sec:algorithm}
Based on the analysis presented in the previous section,
we develop a novel homotopy algorithm for robust SVM.
We call the proposed method the \emph{outlier-path (OP)} algorithm.
For simplicity,
we consider homotopy path computation involving either 
$\theta$
or 
$s$,
and denote the former as 
OP-$\theta$
and the latter as 
OP-$s$.
OP-$\theta$ computes the local solution path when
$\theta$
is gradually decreased from 1 to 0 
with fixed $s = 0$,
while 
OP-$s$ computes the local solution path when
$s$
is gradually increased from $-\infty$ to 0
with fixed $\theta = 0$.

The local optimality of robust SVM in the previous section shows that 
the path of local optimal solutions has finite discontinuous points that satisfy (\ref{eq:local.optimality.condition.f}) or (\ref{eq:local.optimality.condition.f.rsvr}).
Below, we introduce an algorithm that appropriately handles those discontinuous points.
In this section,
we only describe the algorithm for robust SVC.
All the methodologies
described
in this section
can be easily extended to robust SVR counterpart.

\subsection{Overview}
\begin{algorithm}[t]
  \caption{Outlier Path Algorithm}
  \label{alg:OP_alg}
\begin{algorithmic}[1]
 \STATE Initialize the solution $f$ by solving the standard SVM.
 \STATE Initialize the partition $\cP := \{\cI, \cO\}$ as follows:
\begin{small}
 \begin{align*}
  \cI~&\leftarrow~\{i \in \NN_n | y_i f(\bm x_i) \le s\}, \\
  \cO~&\leftarrow~\{i \in \NN_n | y_i f(\bm x_i) > s\}.
 \end{align*}
\end{small}
\vspace*{-5mm}
\STATE $\theta \leftarrow 1$ for OP-$\theta$; $s \leftarrow \min_{i \in \NN_n} y_i f(\bm x_i)$ for OP-$s$. 
 \WHILE{$\theta > 0$ for OP-$\theta$; $s < 0$ for OP-$s$}
\IF{($y_i f(\bm x_i) \neq s ~\forall~ i \in \NN_n$)}
\STATE Run C-step.
\ELSE 
     \STATE Run D-step.
\ENDIF
\ENDWHILE
\end{algorithmic}
\end{algorithm}

The main flow of the OP algorithm is described in Algorithm~\ref{alg:OP_alg}. 
The solution $f$ is initialized by solving the standard (convex) SVM, 
and the partition
$\cP := \{\cI, \cO\}$
is defined to satisfy the constraints in 
\eq{eq:polytope}.
The algorithm mainly switches over the two steps called 
the \emph{continuous step (C-step)} 
and 
the \emph{discontinuous step (D-step)}. 

In the C-step (Algorithm~\ref{alg:C-step}),
a continuous path of local solutions is computed
for a sequence of gradually decreasing
$\theta$ (or increasing $s$)
within the convex polytope
${\rm pol}(\cP;s)$
defined by the current partition $\cP$.
If the local solution path encounters a boundary of the convex polytope,
i.e.,
if there exists at least an instance such that
$y_i f(\bm x_i) = s$,
then the algorithm stops updating
$\theta$ (or $s$)
and enters the D-step.

In the D-step (Algorithm~\ref{alg:D-step}),
a better local solution is obtained for fixed
$\theta$ (or $s$)
by solving a convex problem defined over another convex polytope
in the opposite side of the boundary
(see \figurename~\ref{fig:solution_d}).
If the new solution is again at a boundary of the new polytope,
the algorithm repeatedly calls the D-step
until it finds the solution in the strict interior of the current polytope.

The C-step can be implemented by any homotopy algorithms
for solving a sequence of quadratic problems (QP). 
In OP-$\theta$,
the local solution path can be exactly computed
because the path within a convex polytope can be represented as piecewise-linear functions of the homotopy parameter $\theta$. 
In OP-$s$,
the C-step is trivial because the optimal solution is shown to be constant within a convex polytope. 
In
\S~\ref{subsec:C-step-OP-theta}
and
\S~\ref{subsec:C-step-OP-s}, 
we will describe the details of our implementation of the C-step
for OP-$\theta$ and OP-$s$, respectively. 

\begin{figure}[!h]
 \centering
 \includegraphics[width=0.65\linewidth]{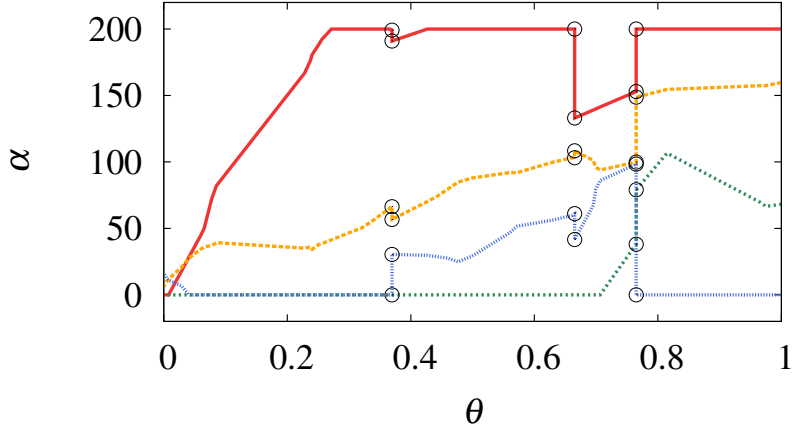}
 \caption{
 An example of the local solution path by OP-$\theta$ on a simple toy data set (with $C = 200$). 
 The paths of five Lagrange multipliers 
 $\alpha^*_1, \cdots, \alpha^*_4$
 are plotted in the range of 
 $\theta \in [0, 1]$. 
 Open circles represent the discontinuous points in the path.
 In this simple example,
 we had experienced three discontinuous points at
 $\theta = 0.37, 0.67$ and $0.77$.
 }
 \label{fig:path.illustration}
\end{figure}

In the D-step, we only need to solve a single quadratic problem (QP). 
Any QP solver can be used in this step.
We note that the \emph{warm-start} approach
\cite{decoste00a}
is quite helpful in the D-step
because the difference between two conditionally optimal solutions in adjacent two convex polytopes is typically very small. 
In
\S~\ref{subsec:D-step},
we describe the details of our implementation of the D-step.
\figurename~\ref{fig:path.illustration}
illustrates an example of the local solution path obtained by OP-$\theta$. 

\begin{algorithm}[t]
  \caption{Continuous Step (C-step)}
  \label{alg:C-step}
\begin{algorithmic}[1]
 \WHILE{($y_i f(\bm x_i) \neq s ~\forall~ i \in \NN_n$)}
 \STATE 
Solve the sequence of convex problems,
\begin{align*}
 \min_{f \in {\rm pol}(\cP;s)} J_{\cP}(f;\theta),
\end{align*}
for gradually decreasing $\theta$ in OP-$\theta$
or gradually increasing $s$ in OP-$s$.
 \ENDWHILE
\end{algorithmic}
\end{algorithm}

\begin{algorithm}[t]
  \caption{Discontinuous Step (D-step)}
  \label{alg:D-step}
\begin{algorithmic}[1]
  \STATE Update the partition $\cP := \{\cI, \cO\}$ as follows:
 \begin{small}
 \begin{align*}
  \cI \; \leftarrow \cI &\setminus \{ i \in \cI | y_i f(\Vec{x}_i) = s \} \cup \{ i \in \cO | y_i f(\Vec{x}_i) = s \}, \\
  \cO \leftarrow \cO &\setminus \{ i \in \cO | y_i f(\Vec{x}_i) = s \} \cup \{ i \in \cI | y_i f(\Vec{x}_i) = s \}.
 \end{align*}
 \end{small}
 \vspace*{-5mm}
 \STATE
 Solve the following convex problem for fixed $\theta$ and $s$:
\begin{align*}
 \min_{f \in {\rm pol}(\cP;s)} J_{\cP}(f;\theta).
\end{align*}
\end{algorithmic}
\end{algorithm}

\subsection{Continuous-Step for OP-$\theta$}
\label{subsec:C-step-OP-theta}
In the C-step,
the partition
$\cP := \{\cI, \cO\}$
is fixed,
and our task is to solve a sequence of convex quadratic problems (QPs)
parameterized by
$\theta$
within the convex polytope
${\rm pol}(\cP;s)$. 
It has been known in optimization literature that
a certain class of parametric convex QP can be exactly solved
by exploiting the piecewise linearity of the solution path
\cite{Best96}.
We can easily show that the local solution path of OP-$\theta$ within a convex polytope 
is also represented as a piecewise-linear function of $\theta$.
The algorithm presented here is similar to 
the regularization path algorithm for SVM given in \cite{HasRosTibZhu04}.

Let us consider a partition of the inliers in $\cI$ into the following three disjoint sets:
\begin{eqnarray*}
 \label{eq:active_set_LER}
 \mathcal{R} &:=& \{ i | 1 < y_i f(\Vec{x}_i) \}, \\
 \mathcal{E} &:=& \{ i | y_i f(\Vec{x}_i) = 1 \}, \\
 \mathcal{L} &:=& \{ i | s < y_i f(\Vec{x}_i) < 1 \}.
\end{eqnarray*}

For a given fixed partition 
$\{\cR, \cE, \cL, \cO\}$,
the KKT conditions of the convex problem
\eq{eq:conditionally.optimal.solution}
indicate that 
\begin{align*}
 \alpha_i = 0
 ~\forall~
 i \in \cR, 
 ~~
 \alpha_i = C
 ~\forall~
 i \in \cL, 
 ~~
 \alpha_i = C \theta
 ~\forall~
i \in \cO.
\end{align*}
The KKT conditions also imply that  
the remaining Lagrange multipliers
$\{\alpha_i\}_{i \in \cE}$
must satisfy the following linear system of equations:
\begin{align}
 \nonumber
 &
 ~~
 y_i f(\bm x_i) = \sum_{j \in \NN_n} \alpha_j y_i y_j K(\bm x_i, \bm x_j) = 1 ~\forall~ i \in \cE
 \\
 \label{eq:linear.system}
 &
 \Leftrightarrow
 ~~
 \bm Q_{\cE \cE}
 \bm \alpha_{\cE}
 =
 \one - 
 \bm Q_{\cE \cL}
 \one
 C
 -
 \bm Q_{\cE \cO}
 \one
 C \theta,
\end{align}
where
$\bm Q \in \RR^{n \times n}$
is an $n \times n$ matrix 
whose
$(i, j)^{\rm th}$
entry is defined as 
$Q_{ij} := y_i y_j K(\Vec{x}_i, \Vec{x}_j)$.
Here,
a notation such as 
$\bm Q_{\cE \cL}$
represents a submatrix of
$\bm Q$
having only the rows in the index set
$\cE$
and the columns in the index set
$\cL$.
By solving the linear system of equations
\eq{eq:linear.system},
the Lagrange multipliers
$\alpha_i, i \in \NN_n$,
can be written as an affine function of $\theta$. 

Noting that
$y_i f(\bm x_i) = \sum_{j \in \NN_n} \alpha_j y_i y_j K(\bm x_i, \bm x_j)$
is also represented as an affine function of $\theta$, 
any changes of the partition
$\{\cR, \cE, \cL\}$
can be exactly identified when the homotopy parameter $\theta$ is continuously decreased. 
Since the solution path linearly changes for each partition of
$\{\cR, \cE, \cL\}$,
the entire path is represented as a continuous piecewise-linear function of the homotopy parameter $\theta$.
We denote the points in $\theta \in [0, 1)$
at which members of the sets
$\{\cR, \cE, \cL\}$
change as
\emph{break-points}
$\theta_{BP}$.
%

Using the piecewise-linearity of
$y_i f(\bm x_i)$,
we can also identify when we should switch to the D-step.
Once we detect an instance satisfying
$y_i f(\bm x_i) = s$,
we exit the C-step and enter the D-step. 

\subsection{Continuous-Step for OP-$s$}
\label{subsec:C-step-OP-s}
Since $\theta$ is fixed to 0 in OP-$s$,
the KKT conditions \eq{eq:KKT_conds} yields
\begin{align*}
 \alpha_i = 0 ~ \forall ~ i \in \cO.
\end{align*}
This means that
outliers have no influence on the solution and thus the conditionally optimal solution
$f^*_{\cP}$
does not change with $s$ 
as long as the partition
$\cP$
is unchanged.
The only task in the C-step for OP-$s$ is
therefore to find the next $s$ that changes the partition $\cP$.
Such $s$ can be simply found as
\begin{align*}
 s ~\leftarrow~ \min_{i \in \cL} y_i f(\bm x_i). 
\end{align*}

\subsection{Discontinuous-Step (for both OP-$\theta$ and OP-$s$)}
\label{subsec:D-step}
As mentioned before,
any convex QP solver can be used for the D-step.
When the algorithm enters the D-step,
we have the conditionally optimal solution
$f^*_\cP$
for the partition
$\cP := \{\cI, \cO\}$.
Our task here is to find another conditionally optimal solution
$f^*_{\tilde{\cP}}$
for 
$\tilde{\cP} := \{\tilde{\cI}, \tilde{\cO}\}$
given by 
\eq{eq:new.I.O}.

Given that the difference between the two solutions
$f^*_\cP$
and 
$f^*_{\tilde{\cP}}$
is typically small,
the D-step can be efficiently implemented by a technique
used in the context of incremental learning
\cite{CauPog01}.

Let us define 
\begin{align*}
 \Delta_{\cI \to \cO} &:= \{i \in \cI ~ | ~ y_i f_{\cP}(\bm x_i) = s \},
 \\
 \Delta_{\cO \to \cI} &:= \{i \in \cO ~ | ~ y_i f_{\cP}(\bm x_i) = s \}.
\end{align*}
Then,
we consider the following parameterized problem with parameter $\mu \in [0, 1]$:
\begin{align*}
 f_{\tilde{\cP}}(\bm x_i; \mu)
 :=
 f_{\tilde{\cP}}(\bm x_i)
 + \mu \Delta f_i 
 ~ \forall ~ i \in \NN_n, 
\end{align*}
where
\begin{align*}
 \Delta f_i :=
 y_i
 \mtx{cc}{
 \bm K_{i, \Delta_{\cI \to \cO}} &
 \bm K_{i, \Delta_{\cO \to \cI}} \\
 }
 \mtx{c}{
 \bm \alpha^{\rm (bef)}_{\Delta_{\cI \to \cO}} - \one C \theta \\
 \bm \alpha^{\rm (bef)}_{\Delta_{\cO \to \cI}} - \one C
 },
\end{align*}
and
$\bm \alpha^{(\rm bef)}$
be the corresponding
$\bm \alpha$
at the beginning of the D-Step.
We can show that 
$f_{\tilde{\cP}}(\bm x_i; \mu)$
is reduced to 
$f_{\cP}(\bm x_i)$
when $\mu = 1$,
while 
it is reduced to 
$f_{\tilde{\cP}}(\bm x_i)$
when $\mu = 0$
for all
$i \in \NN_n$.
By using a similar technique to incremental learning
\cite{CauPog01},
we can efficiently compute the path of solutions
when
$\mu$
is continuously changed from 1 to 0.
This algorithm behaves similarly to the C-step in OP-$\theta$.
The implementation detail of the D-step is described in Appendix \ref{app:Implementation.D-step}.

\section{Numerical Experiments}
\label{sec:experiment}
In this section, 
we compare the proposed outlier-path (OP) algorithm 
with conventional concave-convex procedure (CCCP)
\cite{yuille02a} 
because, 
in most of the existing robust SVM studies,
non-convex optimization for robust SVM training
are solved by 
CCCP or a variant called difference of convex (DC) programming
\cite{Shen03,krause04a,Liu05a,Liu06a,Collobert06a,wu07a}.

\subsection{\bf Setup}
We used several benchmark data sets listed in
Tables
\ref{tb:bench_data_rsvc}
and
\ref{tb:bench_data_rsvr}.
We randomly divided data set into 
training (40\%),
validation (30\%),
and
test (30\%) sets
for the purposes of optimization,
model selection (including the selection of $\theta$ or $s$),
and performance evaluation,
respectively.
For robust SVC,
we randomly flipped 15\% of the labels in the training and the validation data sets.
For robust SVR, 
we first preprocess the input and output variables; 
each input variable was normalized 
so that the minimum and the maximum values are
$-1$ and $+1$,
respectively,
while 
the output variable 
was standardized to have mean zero and variance one. 
Then, 
for the 5\% of the training and the validation instances, 
we added an uniform noise
$U(-2, 2)$
to input variable,
and
a Gaussian noise 
$N(0, 10^2)$
to output variable, where $U(a,b)$ denotes the uniform distribution between $a$ and $b$
and $N(\mu,\sigma^2)$ denotes the normal distribution with mean $\mu$ and variance $\sigma^2$.

\begin{table}[!h]
  \footnotesize
  \centering
  \caption{Benchmark data sets for robust SVC experiments}
  \label{tb:bench_data_rsvc}
  \vspace*{2mm}
  \begin{tabular}{cl|c|c} \hline
	& ~~~~~~Data & $n$ & $d$ \\ \hline
	D1	&	BreastCancerDiagnostic	&	569	&	30	\\
	D2	&	AustralianCredit	&	690	&	14	\\
	D3	&	GermanNumer	&	1000	&	24	\\
	D4	&	SVMGuide1	&	3089 &	4	\\
	D5	&	spambase	&	4601	&	57	\\
	D6	&	musk	&	6598	&	166	\\
	D7	&	gisette	&	6000	&	5000	\\
	D8	&	w5a	&	9888	&	300	\\
	D9	&	a6a	&	11220	&	122	\\
	D10	&	a7a	&	16100	&	122	\\ \hline
   \multicolumn{3}{l}{
   $n = \#$ of instances, $d = $ input dimension} \\
  \end{tabular}
  \caption{Benchmark data sets for robust SVR experiments}
  \label{tb:bench_data_rsvr}
  \begin{tabular}{cl|c|c} \hline
	& ~~~~~~Data & $n$ & $d$ \\ \hline
	D1 &	bodyfat	& 252 & 14 \\
	D2 &	yacht\_hydrodynamics	& 308 & 6 \\
	D3 &	mpg	& 392 & 7 \\
	D4 &	housing	& 506 & 13 \\
	D5 &	mg	& 1385 & 6 \\
	D6 &	winequality-red	& 1599 & 11 \\
	D7 &	winequality-white	& 4898 & 11 \\
	D8 &	space\_ga	& 3107 & 6 \\
	D9 &	abalone	& 4177 & 8 \\
	D10 &	cpusmall	& 8192 & 12 \\
	D11 &	cadata	& 20640 & 8 \\ \hline
   \multicolumn{3}{l}{
   $n = \#$ of instances, $d = $ input dimension} \\
  \end{tabular}
\end{table}

\subsection{Generalization Performance}
First,
we compared the generalization performance.
We used the linear kernel and the radial basis function (RBF) kernel 
defined as
$K(\Vec{x}_i, \Vec{x}_j) = \exp \left( -\gamma \|\Vec{x}_i - \Vec{x}_j\|^2 \right)$, 
where $\gamma$ is a kernel parameter fixed to 
$\gamma=1/d$ with $d$ being the input dimensionality.
Model selection was carried out
by finding the best hyper-parameter combination that minimizes the validation error.
We have a pair of hyper-parameters in each setup.
In all the setups,
the regularization parameter
$C$
was chosen from 
$\{0.01, 0.1, 1, 10, 100\}$,
while 
the candidates of the homotopy parameters
were chosen as follows:
\begin{itemize}
 \item In OP-$\theta$, 
       the set of break-points
       $\theta_{BP} \in [0,1]$
       was considered as the candidates
       (note that the local solutions at each break-point have been already computed in the homotopy computation).
       
 \item In OP-$s$, 
       the set of break-points in 
       $[s_C, 0]$ 
       was used as the candidates for robust SVC, 
       where 
       \begin{align*}
	s_C := \min_{i \in \NN_n} y_i f_{\rm SVC}(\Vec{x}_i)
       \end{align*}
       with
       $f_{\rm SVC}$
       being the ordinary non-robust SVC.
       For robust SVR, 
       the set of break-points in 
       $[s_R, 0.2S_R]$
       was used as the candidates, 
       where
       \begin{align*}
	s_R := \max_{i \in \NN_n} |y_i - f_{\rm SVR}(\Vec{x}_i)|
       \end{align*}
       with
       $f_{\rm SVC}$
       being the ordinary non-robust SVR.
       
 \item In CCCP-$\theta$, 
       the homotopy parameter $\theta$
       was selected from 
       \begin{align*}
	\theta \in \{1, 0.75, 0.5, 0.25, 0\}.
       \end{align*}

 \item In CCCP-$s$,
       the homotopy parameter $s$ was selected from 
       \begin{align*}
	s \in \{s_C, 0.75s_C, 0.5s_C, 0.25s_C, 0\}
       \end{align*}
       for robust SVC,
       while it was selected from
       \begin{align*}
	s \in \{s_R, 0.8s_R, 0.6s_R, 0.4s_R, 0.2s_R\}
       \end{align*}
       for robust SVR.
\end{itemize}
Note that both OP and CCCP were initialized by using the solution of standard SVM. 

Tables
\ref{tb:test_error_linear}-\ref{tb:test_error_rbf_rsvr}
represent the average and the standard deviation of the test errors on 10 different random data splits.
These results indicate that
our proposed OP algorithm tends to find better local solutions and the degree of robustness was appropriately controlled.

\begin{table}[!h]
  \scriptsize
  \centering
  \caption{The mean of test error by 0-1 loss and standard deviation (linear, robust SVC).
  Smaller test error is better. The numbers in bold face indicate the better method in terms of the test error.}
  \label{tb:test_error_linear}
  \vspace*{2mm}
  \begin{tabular}{r||c||c|c||c|c} \hline
	Data	&	$C$-SVC	&	CCCP-$\theta$	&	OP-$\theta$	&	CCCP-$s$	&	OP-$s$	\\ \hline
D1	&	.056(.016)	&	.050(.014)	&	{\bf .049(.016)}	&	.055(.018)	&	{\bf .050(.016)}	 \\ \hline
D2	&	.151(.018)	&	{\bf .145(.007)}	&	.151(.018)	&	{\bf .145(.007)}	&	.152(.010)	 \\ \hline
D3	&	.281(.028)	&	.270(.033)	&	.270(.023)	&	{\bf .262(.013)}	&	.266(.013)	 \\ \hline
D4	&	.066(.007)	&	.047(.007)	&	.047(.005)	&	.053(.010)	&	{\bf .042(.006)}	 \\ \hline
D5	&	.108(.010)	&	.088(.009)	&	.088(.009)	&	.088(.010)	&	{\bf .084(.007)}	 \\ \hline
D6	&	.072(.005)	&	{\bf .058(.006)}	&	.064(.003)	&	.061(.007)	&	{\bf .060(.003)}	 \\ \hline
D7	&	.185(.013)	&	.184(.010)	&	.184(.010)	&	.184(.010)	&	.184(.010)	 \\ \hline
D8	&	.020(.002)	&	.020(.003)	&	.020(.002)	&	.021(.003)	&	{\bf .020(.003)}	 \\ \hline
D9	&	.173(.004)	&	.181(.009)	&	{\bf .173(.005)}	&	.165(.004)	&	{\bf .164(.004)}	 \\ \hline
D10	&	.173(.008)	&	.176(.006)	&	{\bf .173(.007)}	&	{\bf .160(.004)}	&	.161(.005)	 \\ \hline
  \end{tabular}
  \caption{The mean of test error by 0-1 loss and standard deviation (RBF, robust SVC).}
  \label{tb:test_error_rbf}
  \begin{tabular}{r||c||c|c||c|c} \hline
	Data	&	$C$-SVC	&	CCCP-$\theta$	&	OP-$\theta$	&	CCCP-$s$	&	OP-$s$	\\ \hline
D1	&	.055(.017)	&	.043(.022)	&	{\bf .042(.017)}	&	{\bf .037(.016)}	&	.038(.013)	 \\ \hline
D2	&	.149(.010)	&	.148(.010)	&	{\bf .147(.010)}	&	.146(.013)	&	{\bf .142(.013)}	 \\ \hline
D3	&	.276(.024)	&	.267(.026)	&	{\bf .266(.024)}	&	.271(.015)	&	{\bf .261(.020)}	 \\ \hline
D4	&	.052(.009)	&	.048(.009)	&	{\bf .044(.006)}	&	.047(.008)	&	{\bf .040(.005)}	 \\ \hline
D5	&	.117(.012)	&	.109(.013)	&	{\bf .107(.012)}	&	.107(.011)	&	{\bf .094(.008)}	 \\ \hline
D6	&	.046(.007)	&	.045(.007)	&	.045(.007)	&	.045(.007)	&	{\bf .043(.006)}	 \\ \hline
D7	&	.044(.003)	&	.044(.003)	&	.044(.003)	&	.044(.003)	&	.044(.003)	 \\ \hline
D8	&	.022(.003)	&	.022(.003)	&	.022(.003)	&	.022(.003)	&	{\bf .021(.002)}	 \\ \hline
D9	&	.169(.003)	&	.170(.005)	&	{\bf .169(.004)}	&	.168(.005)	&	{\bf .162(.003)}	 \\ \hline
D10	&	.163(.003)	&	.163(.003)	&	.163(.003)	&	.162(.002)	&	{\bf .160(.004)}	 \\ \hline
  \end{tabular}
\end{table}
\begin{table}[!h]
  \scriptsize
  \centering
  \caption{The mean of $L_1$ test error and standard deviation (linear, robust SVR).}
  \label{tb:test_error_linear_rsvr}
  \begin{tabular}{r||c||c|c||c|c} \hline
	Data	&	$C$-SVR	&	CCCP-$\theta$	&	OP-$\theta$	&	CCCP-$s$	&	OP-$s$	\\ \hline
D1	&	.442(.324)	&	.337(.347)	&	{\bf .319(.353)}	&	.414(.341)	&	{\bf .276(.321)}	 \\ \hline
D2	&	.470(.053)	&	.487(.086)	&	{\bf .474(.087)}	&	.490(.108)	&	{\bf .484(.104)}	 \\ \hline
D3	&	.414(.038)	&	.351(.025)	&	{\bf .350(.036)}	&	.414(.105)	&	{\bf .372(.043)}	 \\ \hline
D4	&	.548(.180)	&	.520(.193)	&	{\bf .510(.146)}	&	{\bf .562(.210)}	&	.596(.297)	 \\ \hline
D5	&	.539(.019)	&	.531(.019)	&	{\bf .530(.017)}	&	.539(.024)	&	{\bf .529(.018)}	 \\ \hline
D6	&	.685(.028)	&	.664(.026)	&	{\bf .655(.027)}	&	{\bf .685(.044)}	&	.686(.040)	 \\ \hline
D7	&	.700(.016)	&	.691(.017)	&	{\bf .685(.017)}	&	.698(.022)	&	.{\bf 692(.014)}	 \\ \hline
D8	&	.582(.027)	&	.583(.042)	&	{\bf .570(.031)}	&	.589(.035)	&	{\bf .569(.028)}	 \\ \hline
D9	&	.518(.015)	&	.510(.019)	&	{\bf .501(.021)}	&	.522(.026)	&	{\bf .516(.019)}	 \\ \hline
D10	&	.281(.021)	&	{\bf .278(.016)}	&	.279(.016)	&	.269(.018)	&	.269(.021)	 \\ \hline
D11	&	.494(.010)	&	.488(.011)	&	{\bf .487(.012)}	&	.492(.009)	&	.492(.008)	 \\ \hline
  \end{tabular}
  \caption{The mean of $L_1$ test error and standard deviation (RBF, robust SVR).}
  \label{tb:test_error_rbf_rsvr}
  \begin{tabular}{r||c||c|c||c|c} \hline
	Data	&	$C$-SVR	&	CCCP-$\theta$	&	OP-$\theta$	&	CCCP-$s$	&	OP-$s$	\\ \hline
D1	&	.077(.049)	&	.069(.054)	&	{\bf .065(.056)}	&	.070(.053)	&	{\bf .051(.040)}	 \\ \hline
D2	&	.357(.059)	&	.346(.045)	&	{\bf .339(.045)}	&	.332(.038)	&	{\bf .327(.040)}	 \\ \hline
D3	&	.337(.052)	&	{\bf .299(.021)}	&	.302(.019)	&	.296(.022)	&	{\bf .295(.022)}	 \\ \hline
D4	&	.390(.046)	&	.350(.025)	&	{\bf .349(.023)}	&	.357(.022)	&	{\bf .343(.024)}	 \\ \hline
D5	&	.513(.024)	&	.519(.028)	&	{\bf .504(.018)}	&	.515(.024)	&	{\bf .503(.019)}	 \\ \hline
D6	&	.641(.028)	&	.640(.015)	&	{\bf .635(.017)}	&	.634(.022)	&	{\bf .631(.017)}	 \\ \hline
D7	&	.671(.011)	&	.669(.009)	&	.669(.007)	&	.674(.011)	&	{\bf .671(.009)}	 \\ \hline
D8	&	.528(.027)	&	.504(.027)	&	{\bf .496(.024)}	&	.511(.018)	&	{\bf .510(.020)}	 \\ \hline
D9	&	.488(.012)	&	.490(.016)	&	{\bf .486(.012)}	&	.484(.013)	&	{\bf .482(.014)}	 \\ \hline
D10	&	.198(.015)	&	.198(.027)	&	{\bf .196(.025)}	&	.194(.015)	&	{\bf .189(.017)}	 \\ \hline
D11	&	.456(.016)	&	.441(.005)	&	.441(.006)	&	{\bf .444(.015)}	&	.446(.015)	 \\ \hline
  \end{tabular}
\end{table}

\begin{figure*}[!h]
  \centering
  \subfigure[Elapsed time for CCCP and proposed OP (linear, robust SVC)] {
    \includegraphics[width=0.3\textwidth]{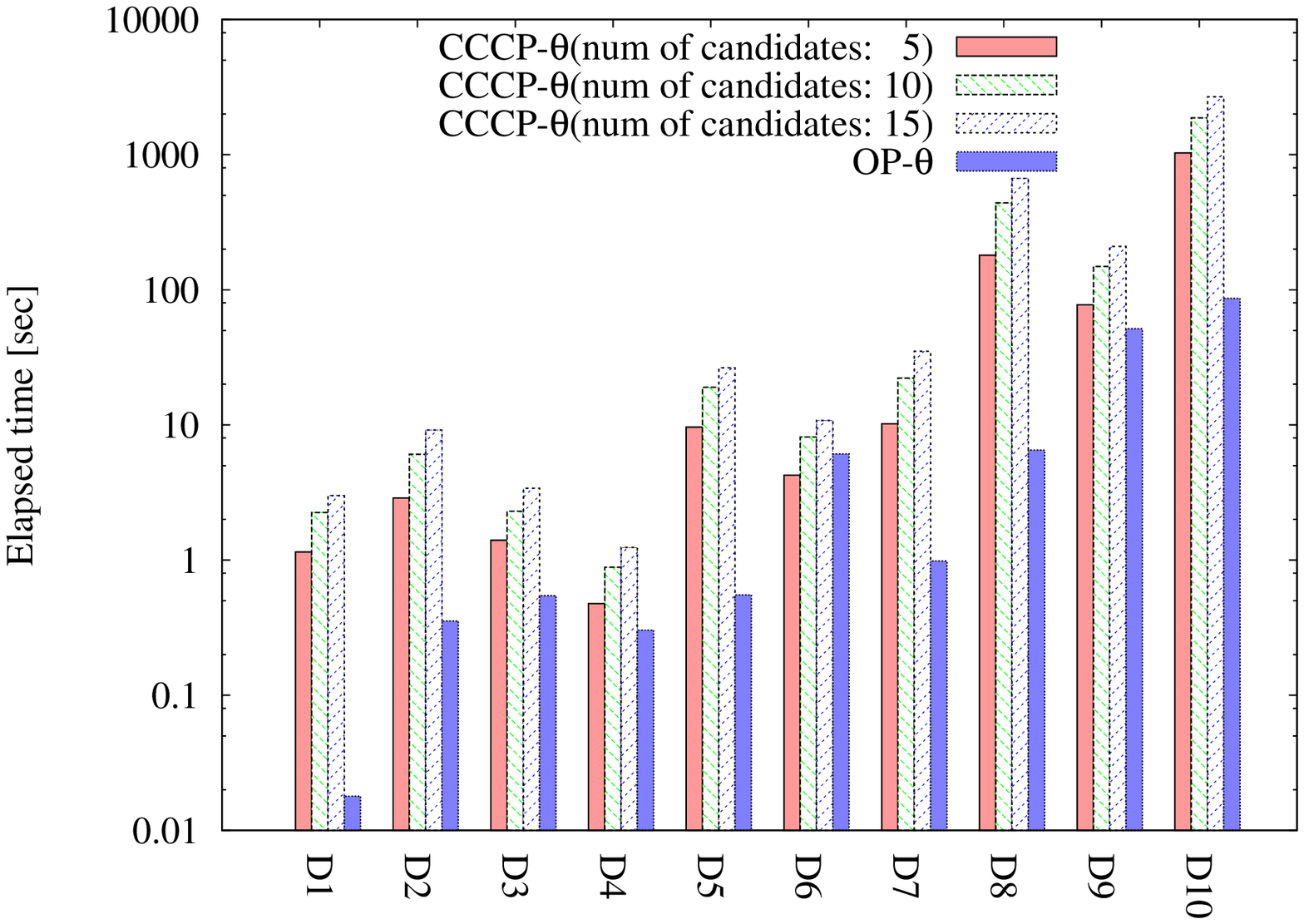}
    \includegraphics[width=0.3\textwidth]{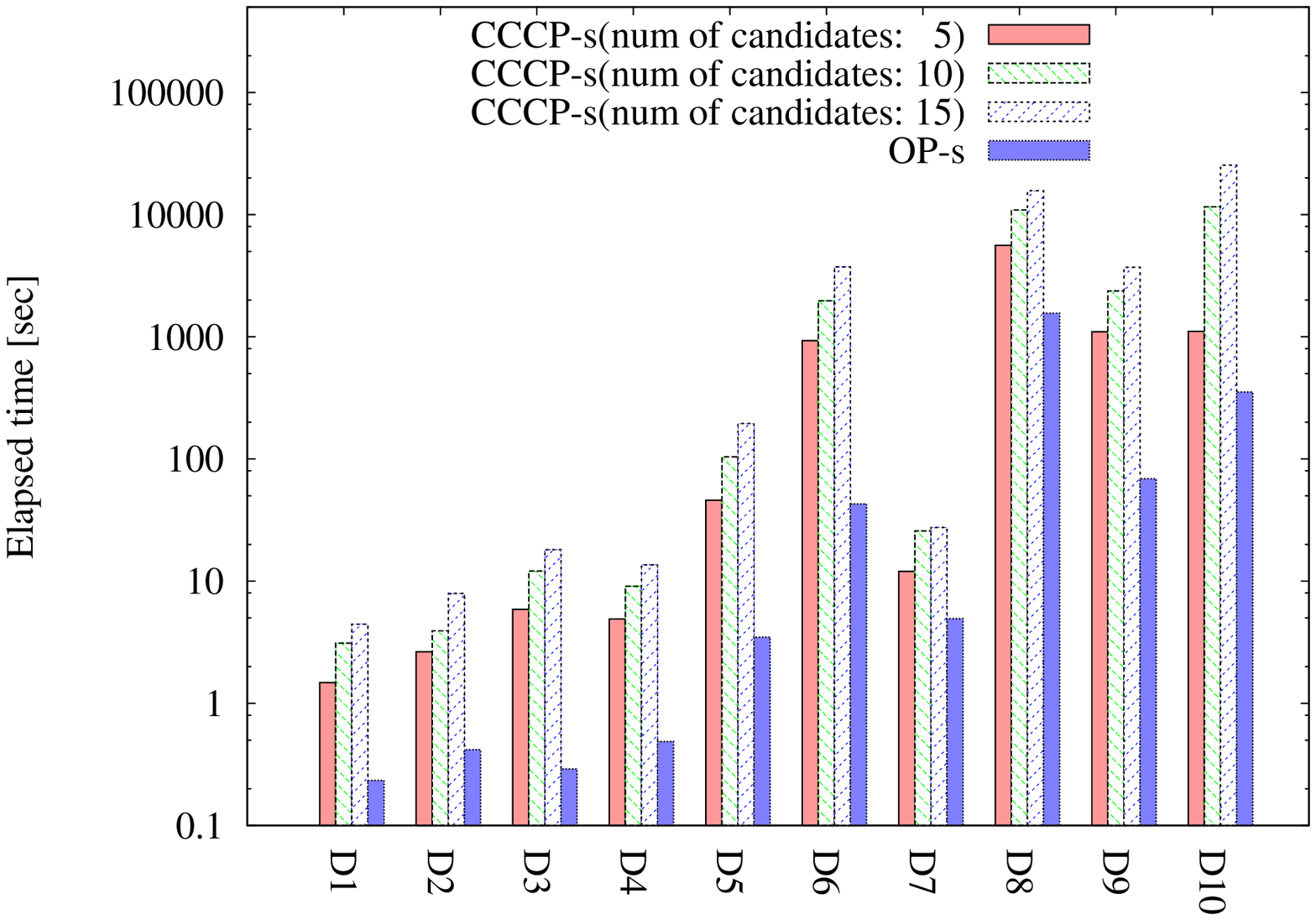}
  }
  \subfigure[Elapsed time for CCCP and proposed OP (RBF, robust SVC)] {
    \includegraphics[width=0.3\textwidth]{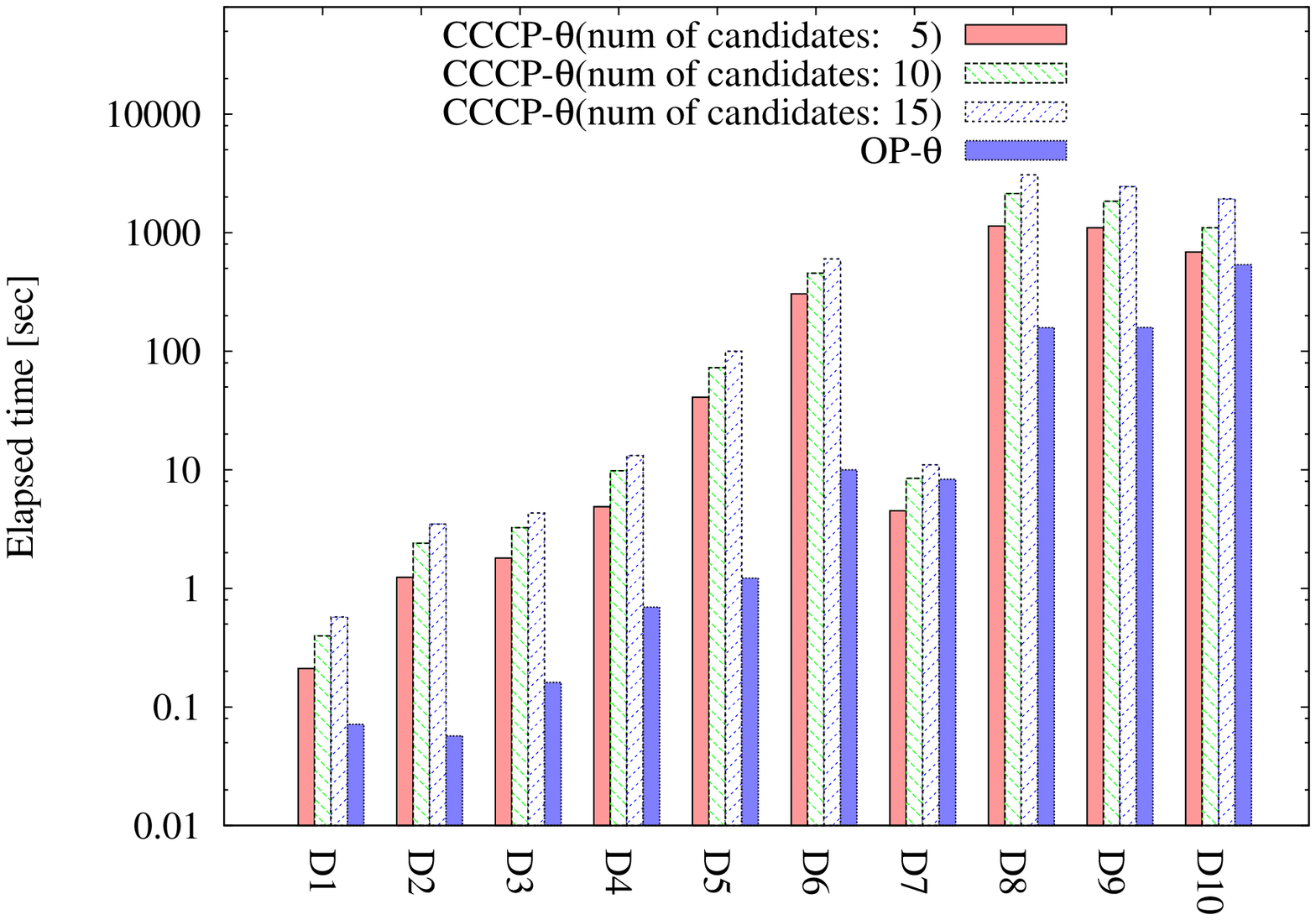}
    \includegraphics[width=0.3\textwidth]{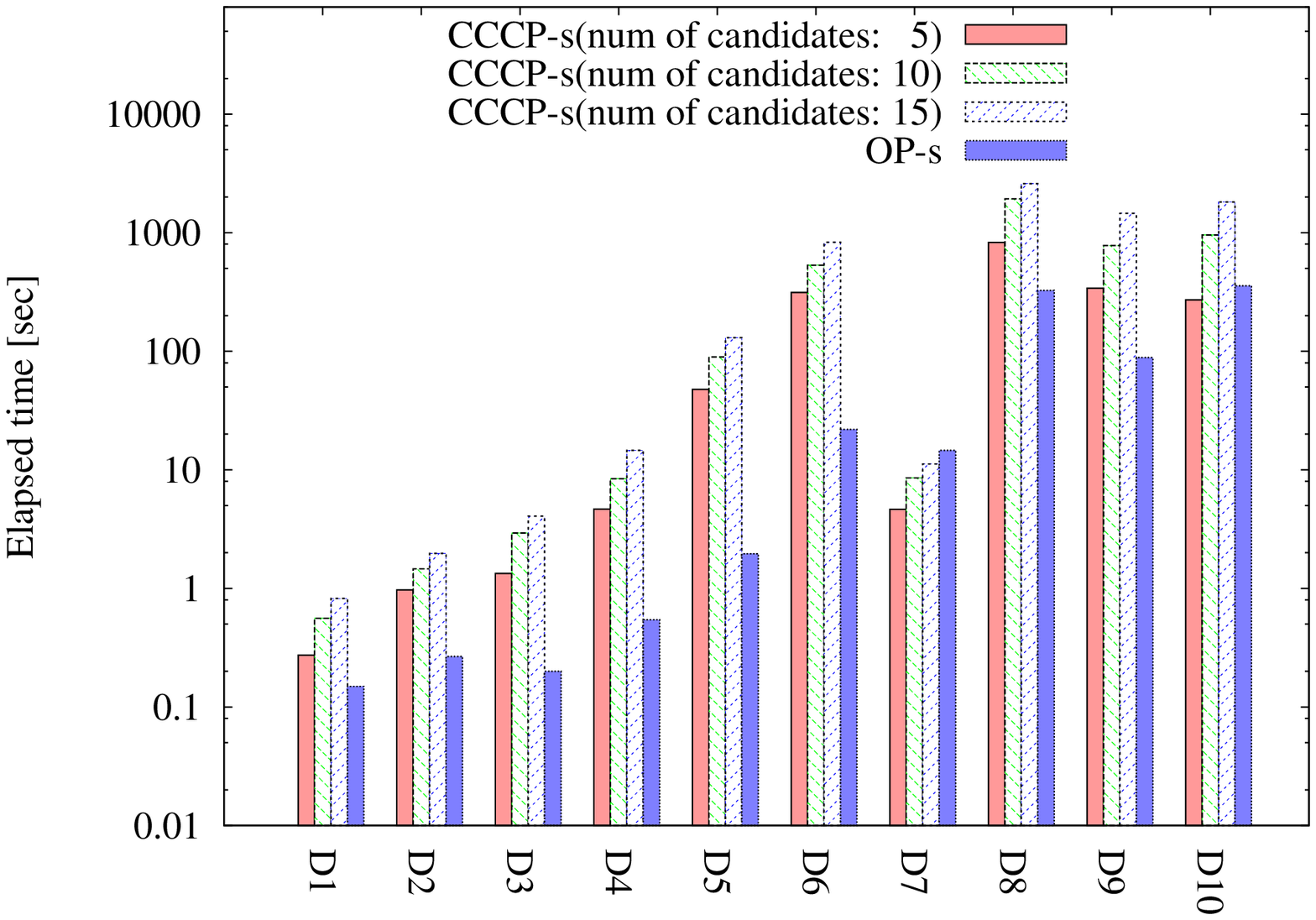}
  }
  \subfigure[Elapsed time for CCCP and proposed OP (linear, robust SVR)] {
    \includegraphics[width=0.3\textwidth]{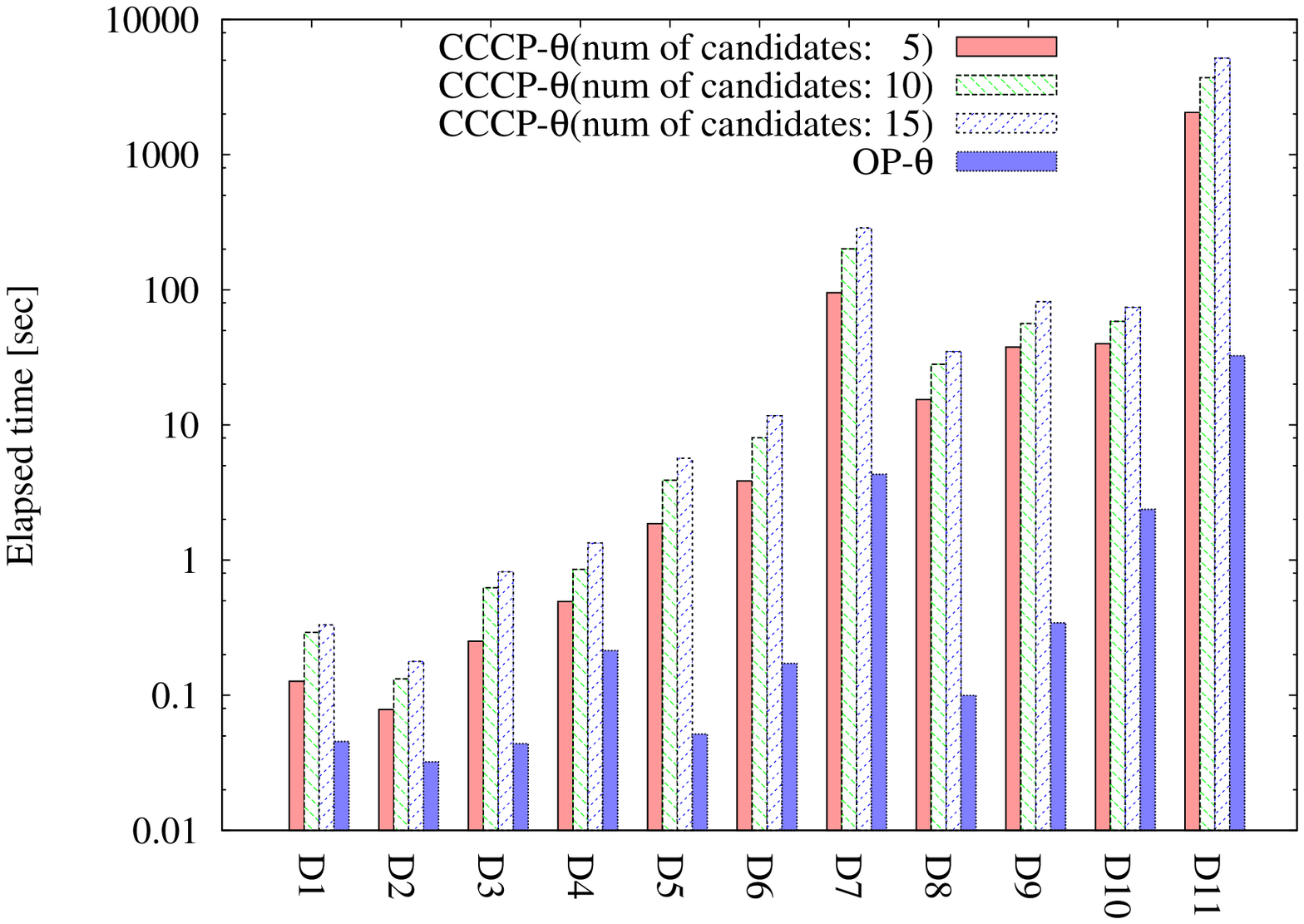}
    \includegraphics[width=0.3\textwidth]{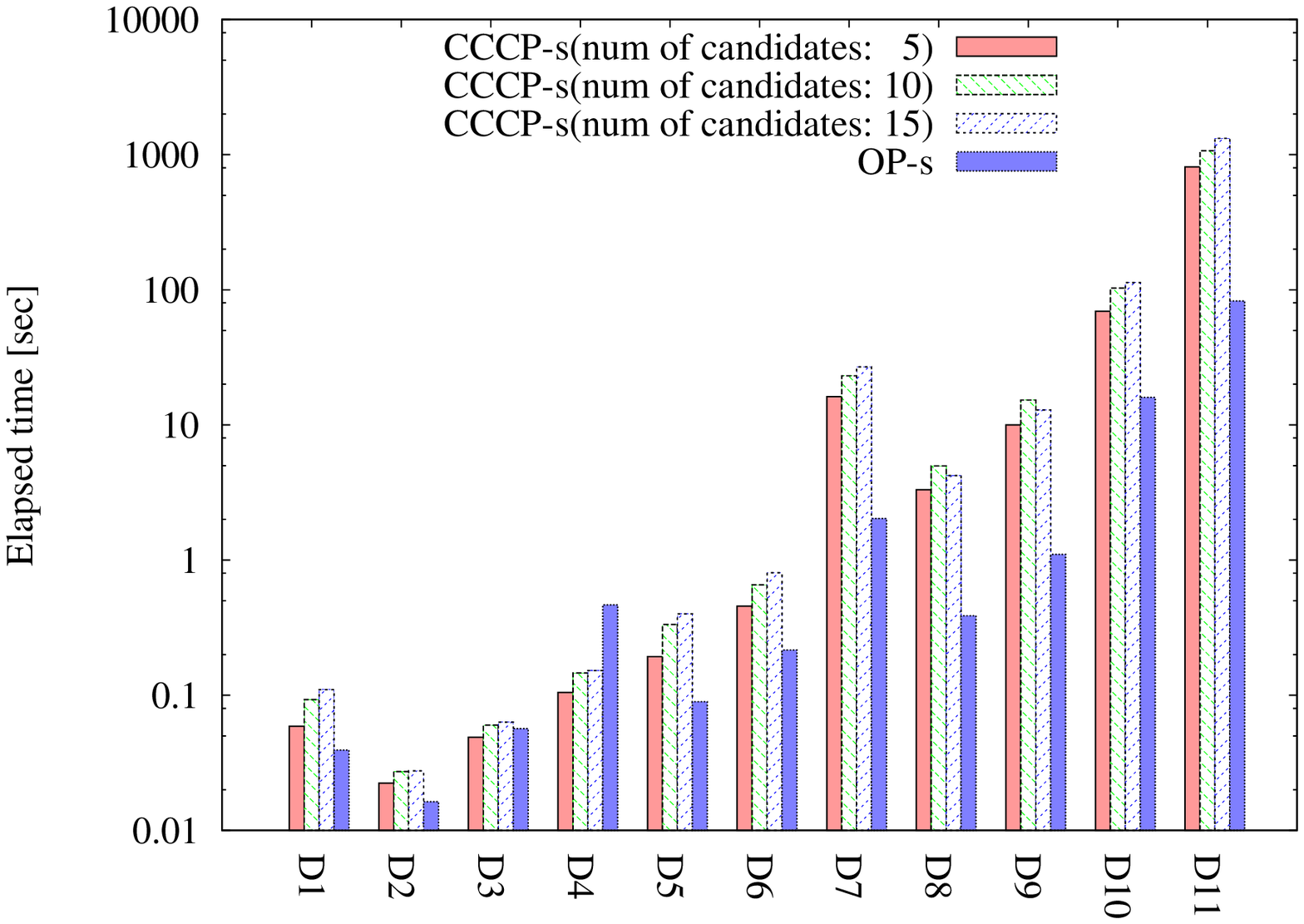}
  }
  \subfigure[Elapsed time for CCCP and proposed OP (RBF, robust SVR)] {
    \includegraphics[width=0.3\textwidth]{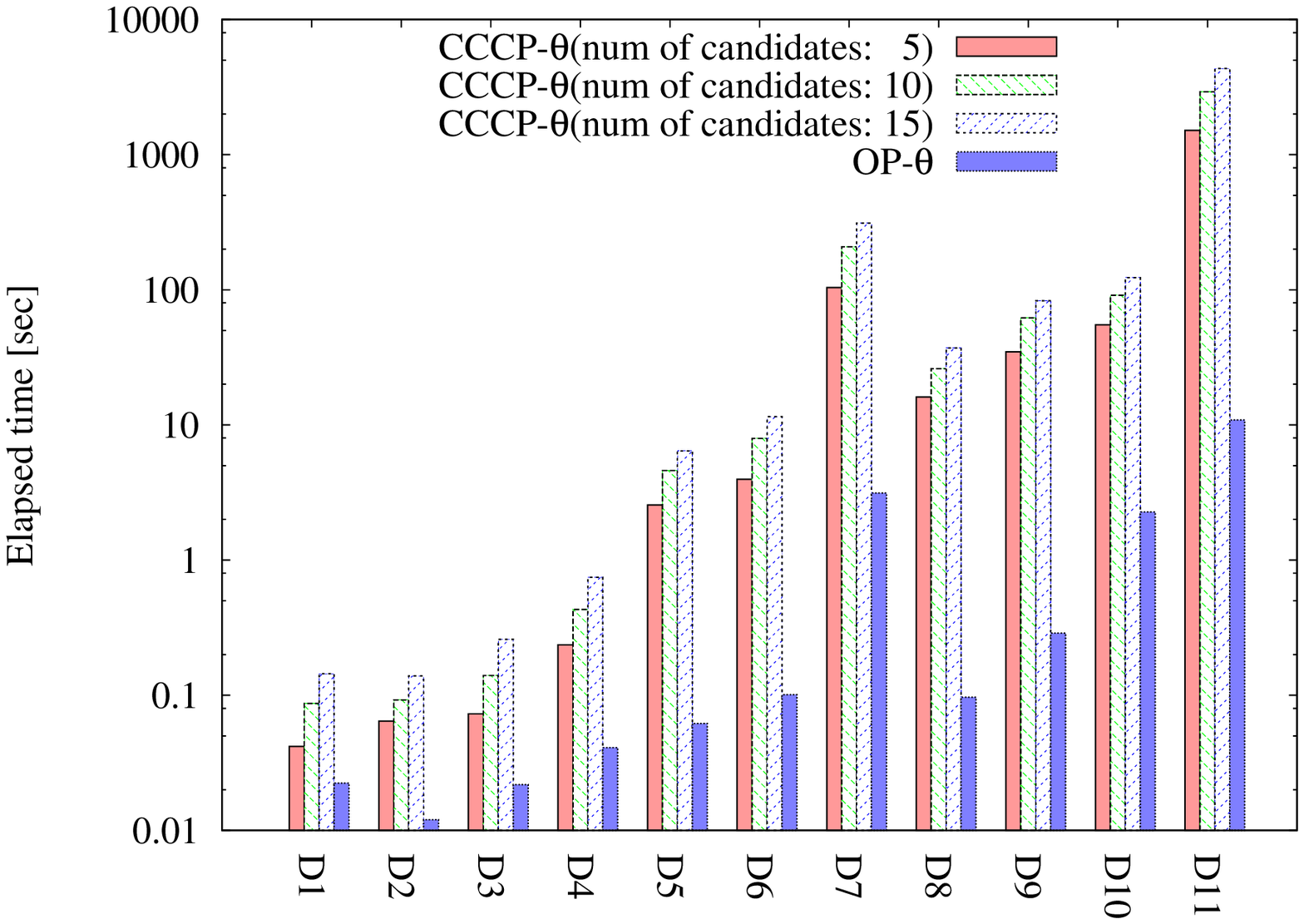}
    \includegraphics[width=0.3\textwidth]{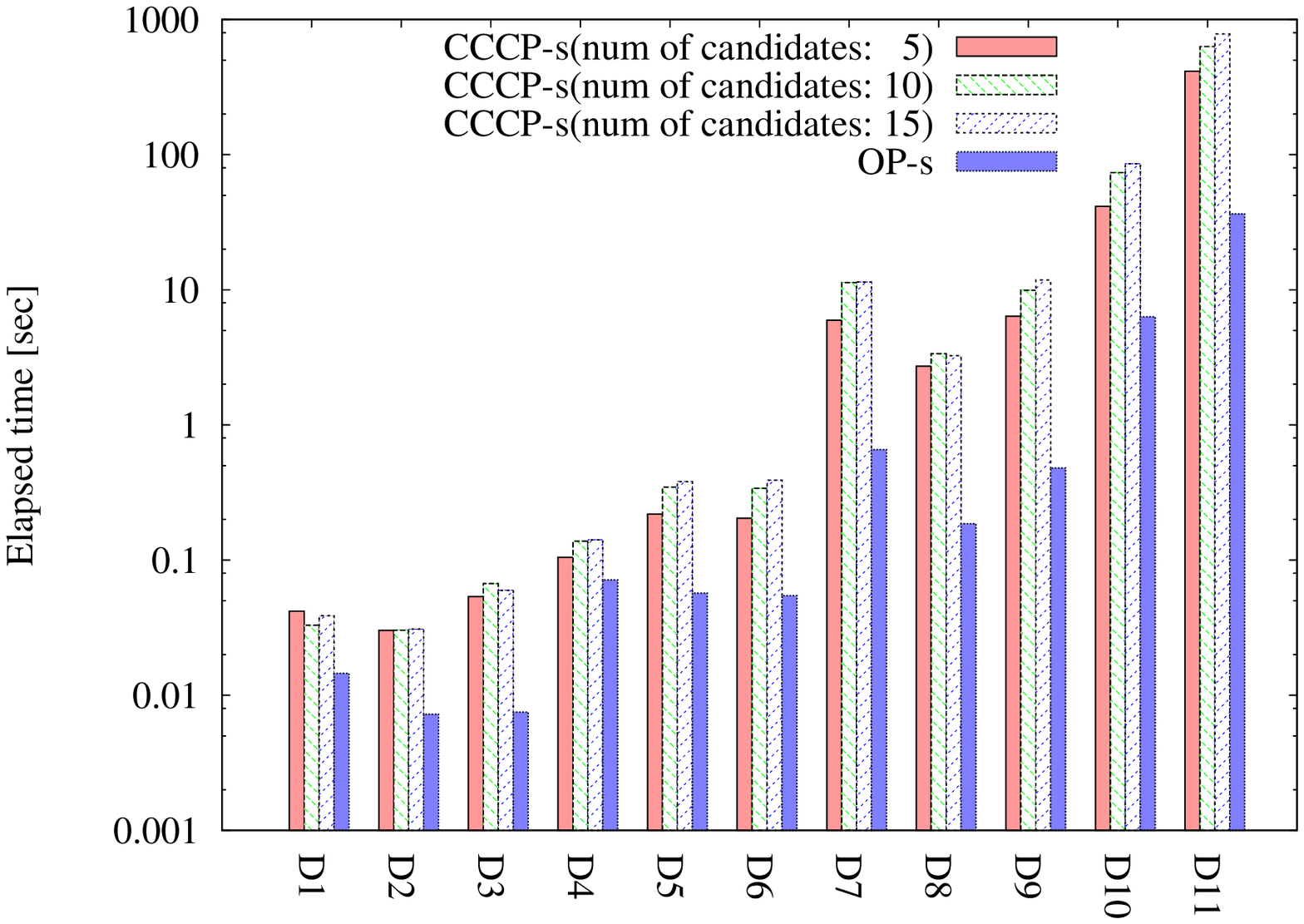}
  }
 \caption{Elapsed time when the number of ($\theta,s$)-candidates is increased.
   Changing the number of hyper-parameter candidates
   affects the computation time of CCCP, but not OP
   because the entire path of solutions is computed with the infinitesimal resolution.
 }
 \label{fig:increased_elapsed_time}
\end{figure*}

\newpage
\subsection{Computation Time}
Finally,
we compared the computational costs of the entire model-building process of each method.
The results are shown in 
\figurename~\ref{fig:increased_elapsed_time}.
Note that the computational cost of the OP algorithm
does not depend on the number of hyper-parameter candidates of 
$\theta$ or $s$,
because the entire path of local solutions has already been computed 
with the infinitesimal resolution in the homotopy computation.
On the other hand,
the computational cost of CCCP depends on the number of hyper-parameter candidates.
In our implementation of CCCP, 
we used the warm-start approach, i.e., we initialized CCCP with the previous solution
for efficiently computing a sequence of solutions.
The results indicate that 
the proposed OP algorithm enables stable and efficient control of robustness,
while
CCCP suffers a trade-off between model selection performance and computational costs.

\newpage
\section{Conclusions}
\label{sec:conclusions}
In this paper, we proposed a novel robust SVM learning algorithm based on the homotopy approach that allows efficient computation of the sequence of local optimal solutions
when the influence of outliers is gradually emphasized.
The algorithm is built on our theoretical findings about the geometric property and the optimality conditions of local solutions of robust SVM. 
Experimental results indicate that our algorithm tends to find better local solutions possibly due to the simulated annealing-like effect and the stable control of robustness.
One of the important future works is to adopt scalable homotopy algorithms \cite{zhou12a} or approximate parametric programming algorithms \cite{Giesen12a} 
for further improving the computational efficiency.

\bibliographystyle{IEEEtran}
\bibliography{eref.bib,suzu_ref.bib}

\clearpage
\appendix
\section{Proof of Theorem~\ref{theo:label.flip}}
\label{app:lemm:strictly.better}
Although $f^*_{\cP}$ is a feasible solution,
it is not a local optimum for $\theta \in [0, 1)$ and $s \le 0$ because
\begin{subequations}
 \label{eq:not.optimal}
\begin{eqnarray}
 \alpha_i \le C\theta &\text{for}& i \in \tilde{\cI} \cap \cO, \\
 \alpha_i \ge C       &\text{for}& i \in \tilde{\cO} \cap \cI,
\end{eqnarray}
\end{subequations}
violate the KKT conditions (\ref{eq:KKT_conds}) for $\tilde{\cP}$.
These feasibility and \emph{sub}-optimality indicates that 
\begin{eqnarray}
 \label{eq:strictly.better.proof.1}
 J_{\tilde{\cP}}(f^*_{\tilde{\cP}}; \theta) < J_{\cP}(f^*_{\cP}; \theta),
\end{eqnarray}
 we arrive at \eq{eq:strictly_better}.
\hfill{\bf Q.E.D.}


\section{Proof of Theorem~\ref{theo:local.optimality.condition}}
\label{app:theo:local.optimality.condition}
{\bf Sufficiency:}
If \eq{eq:local.optimality.condition.f} is true, i.e., if there are NO
instances with $y_i f^*_\cP(\Vec{x}_i) = s$, then any convex problems
defined by different partitions $\tilde{\cP} \neq \cP$ do not have feasible solutions in the neighborhood of $f^*_\cP$.
This means that if $f^*_\cP$ is a conditionally optimal solution, then it is locally optimal.
\eq{eq:local.optimality.condition.a}-\eq{eq:local.optimality.condition.d}
are sufficient for $f^*_\cP$ to be conditionally optimal
for the given partition $\cP$.
Thus, \eq{eq:local.optimality.condition} is sufficient for $f^*_\cP$ to be
locally optimal.

{\bf Necessity:} From Theorem~\ref{theo:label.flip}, if there exists
an instance such that $y_i f^*_\cP(\Vec{x}_i) = s$, then $f^*_\cP$ is a feasible but
not locally optimal.
Then \eq{eq:local.optimality.condition.f} is necessary for $f^*_\cP$ to be locally optimal.
In addition, \eq{eq:local.optimality.condition.a}-\eq{eq:local.optimality.condition.d} are also
necessary for local optimality, because of every local
optimal solutions are conditionally optimal for the given partition $\cP$.
Thus, \eq{eq:local.optimality.condition} is necessary for $f^*_\cP$ to be
locally optimal.

\hfill{\bf Q.E.D.}


\section{Implementation of D-step}
\label{app:Implementation.D-step}
In D-step, we work with the following convex problem
\begin{align}
 f^*_{\tilde{\cP}} := \argmin_{f \in {\rm pol}(\tilde{\cP};s)} J_{\tilde{\cP}}(f;\theta).
\end{align}
where, $\tilde{\cP}$ is updated from $\cP$ as (\ref{eq:new.I.O}).

Let us define a partition 
$\Pi := \{\cR, \cE, \cL, \tilde{\cI}', \tilde{\cO}', \hat{\cO}''\}$
of 
$\NN_n$
such that 
\begin{subequations}
\label{eq:RELIOO}
\begin{eqnarray}
\label{eq:RELIOO.a}
 i \in \cR &\Rightarrow& y_i f(\bm{x}_i) > 1, \\
\label{eq:RELIOO.b}
 i \in \cE &\Rightarrow& y_i f(\bm{x}_i) = 1, \\
\label{eq:RELIOO.c}
 i \in \cL &\Rightarrow& s < y_i f(\bm{x}_i) < 1, \\
\label{eq:RELIOO.d}
 i \in \tilde{\cI}^\prime &\Rightarrow& y_i f(\bm{x}_i) = s \text{ and } i \in \tilde{\cI}, \\
\label{eq:RELIOO.e}
 i \in \tilde{\cO}^\prime &\Rightarrow& y_i f(\bm{x}_i) = s \text{ and } i \in \tilde{\cO}, \\ 
\label{eq:RELIOO.f}
 i \in \tilde{\cO}'' &\Rightarrow& y_i f(\bm{x}_i) < s.
\end{eqnarray}
\end{subequations}

If we write the conditionally optimal solution as 
\begin{eqnarray}
 f^*_{\tilde{\cP}}(x) := \sum_{j \in \NN_n} \alpha_j^* y_j K(x, \bm{x}_j),
\end{eqnarray}
$\{\alpha^*_j\}_{j \in \NN_n}$ 
must satisfy the following KKT conditions
\begin{subequations}
\begin{align}
 y_i f^*_{\tilde{\cP}}(\Vec{x}_i) > 1 ~ & \Rightarrow ~ \alpha_i^* = 0 \\
 y_i f^*_{\tilde{\cP}}(\Vec{x}_i) = 1 ~ & \Rightarrow ~ \alpha_i^* \in [0, C], \\
 s < y_i f^*_{\tilde{\cP}}(\Vec{x}_i) < 1 ~ & \Rightarrow ~ \alpha_i^* = C \\
 y_i f^*_{\tilde{\cP}}(\Vec{x}_i) = s, i \in \tilde{\cI}' ~ & \Rightarrow ~ \alpha_i^* \ge C, \\
 y_i f^*_{\tilde{\cP}}(\Vec{x}_i) = s, i \in \tilde{\cO}' ~ & \Rightarrow ~ \alpha_i^* \le C \theta, \\
 y_i f^*_{\tilde{\cP}}(\Vec{x}_i) < s, i \in \tilde{\cO}'' ~ & \Rightarrow ~ \alpha_i^* = C \theta.
\end{align}
\end{subequations}

At the beginning of the D-step, 
$f^*_{\tilde{\cP}}(\bm{x}_i)$
violates the KKT conditions by 
\begin{align*}
 \Delta f_i :=
 y_i
 \mtx{cc}{
 \bm K_{i, \Delta_{\cI \to \cO}} &
 \bm K_{i, \Delta_{\cO \to \cI}} \\
 }
 \mtx{c}{
 \bm \alpha^{\rm (bef)}_{\Delta_{\cI \to \cO}} - \one C \theta \\
 \bm \alpha^{\rm (bef)}_{\Delta_{\cO \to \cI}} - \one C
 }.
\end{align*}
where $\bm \alpha^{(\rm bef)}$ 
is the corresponding $\bm{\alpha}$ at the beginning of the D-step, 
while 
$\Delta_{\cI \to \cO}$ and $\Delta_{\cO \to \cI}$ denote the difference in $\tilde{\cP}$ and $\cP$ defined as
\begin{align*}
 \Delta_{\cI \to \cO} &:= \{i \in \cI ~ | ~ y_i f_{\cP}(\bm x_i) = s \},
 \\
 \Delta_{\cO \to \cI} &:= \{i \in \cO ~ | ~ y_i f_{\cP}(\bm x_i) = s \}.
\end{align*}

Then,
we consider the following another parametrized problem with a parameter $\mu \in [0, 1]$:
\begin{align*}
 f_{\tilde{\cP}}(\bm x_i; \mu)
 :=
 f_{\tilde{\cP}}(\bm x_i)
 + \mu \Delta f_i 
 ~ \forall ~ i \in \NN_n.
\end{align*}

In order to always satisfy the KKT conditions for $f_{\tilde{\cP}}(\bm x_i; \mu)$, 
we solve the following linear system
\begin{align*}
 \bm{Q}_{\cA, \cA}
 \mtx{c}{
 \bm \alpha_{\cE} \\
 \bm \alpha_{\tilde{\cI}'} \\
 \bm \alpha_{\tilde{\cO}'}
 }
 =
 \mtx{c}{
 \bm 1 \\
 \bm s \\
 \bm s
 } - \bm{Q}_{\cA, \cL} \one C - \bm{Q}_{\cA, \tilde{\cO}''} \one C \theta \\
 -
 \mtx{cc}{
 \bm Q_{\cA, \Delta_{\cI \to \cO}} &
 \bm Q_{\cA, \Delta_{\cO \to \cI}} \\
 }
 \mtx{c}{
 \bm \alpha^{\rm (bef)}_{\Delta_{\cI \to \cO}} - \one C \theta \\
 \bm \alpha^{\rm (bef)}_{\Delta_{\cO \to \cI}} - \one C
 } \mu,
\end{align*}
where $\cA := \{\cE, \tilde{\cI}', \tilde{\cO}'\}$.
This linear system can also be solved by using the piecewise-linear parametric programming while the scalar parameter $\mu$ is continuously moved from 1 to 0.

In this parametric problem, we can show that
$f^*_{\tilde{\cP}}(\bm x_i; \mu) = f^*_{\cP}(\bm x_i)$
 if $\mu = 1$
and 
$f^*_{\tilde{\cP}}(\bm x_i; \mu) = f^*_{\tilde{\cP}}(\bm x_i)$
if $\mu = 0$ for all
$i \in \NN_n$.


Since the number of elements in $\Delta_{\cI \to \cO}$ and $\Delta_{\cO \to \cI}$
 are typically small,
the D-step can be efficiently implemented by a technique
used in the context of incremental learning~\cite{CauPog01}.

\end{document}